\pdfoutput=1
\documentclass{article}

\usepackage{microtype}
\usepackage{calc}
\usepackage{graphicx}
\usepackage{subfigure}
\usepackage{booktabs} %

\usepackage[dvipsnames,table]{xcolor}
\usepackage[export]{adjustbox}
\usepackage{mathtools}
\usepackage{pifont}
\usepackage{microtype}
\usepackage[subtle]{savetrees}
\usepackage{multirow}
\usepackage{makecell}
\usepackage{extarrows}
\usepackage{scrextend}
\usepackage{adjustbox}

\newlength{\itemwidth}

\usepackage{hyperref}

\usepackage[accepted]{icml2023}

\usepackage{amsmath}
\usepackage{amssymb}
\usepackage{mathtools}
\usepackage{amsthm}

\usepackage[
	textsize=tiny,
	textwidth=3.2cm,
	colorinlistoftodos,
	backgroundcolor=teal!30!white,
	linecolor=teal!50!white,
	]{todonotes}

\newcommand{\firstplace}{\cellcolor{gray!35}\boldmath}
\newcommand{\secondplace}{\cellcolor{gray!20}}



\usepackage[capitalize,noabbrev,nameinlink]{cleveref}
\definecolor{ourred}{RGB}{184, 48, 100}
\definecolor{ourblue}{RGB}{100,134,184}

\hypersetup{
	final,
	colorlinks,
	linkcolor=ourred,
	citecolor=ourblue
}
\usepackage{etoolbox}
\newcommand{\projecturl}[1]{{\hypersetup{urlcolor=ourred}\href{#1}{#1}}}
\makeatletter
\patchcmd{\printAffiliationsAndNotice}{Correspondence to: \icmlcorrespondingauthor@text.}
  {Correspondence to: \icmlcorrespondingauthor@text.\\Code: \projecturl{https://github.com/yingheng-wang/InfoDiffusion}}
  {}{\PackageError{main}{could not add the code link to the footnote}{}}
\makeatother

\usepackage{todonotes}
\usetikzlibrary{positioning,arrows.meta,fit,calc}

\theoremstyle{plain}
\newtheorem{theorem}{Theorem}[section]
\newtheorem{proposition}[theorem]{Proposition}

\theoremstyle{definition}

\theoremstyle{remark}

\newcommand{\bx}{\mathbf{x}}
\newcommand{\bz}{\mathbf{z}}
\newcommand{\bs}{\mathbf{s}}
\newcommand{\bh}{\mathbf{h}}
\newcommand{\bb}{\mathbf{b}}

\newcommand{\be}{\mathbf{e}}
\newcommand{\bv}{\mathbf{v}}
\newcommand{\bsigma}{{\boldsymbol{\sigma}}}
\newcommand\Exp{\mathbb{E}}
\newcommand\KL{\text{KL}}
\newcommand\Lc{\mathcal{L}}
\newcommand\Hc{\mathcal{H}}

\newcommand\Rc{\mathcal{R}}
\newcommand\MI{\text{MI}}
\newcommand\D{\text{D}}

\newcommand{\bepsilon}{{\boldsymbol{\epsilon}}}
\newcommand{\bmu}{{\boldsymbol{\mu}}}

\newcommand{\bSigma}{{\boldsymbol{\Sigma}}}

\usepackage[textsize=tiny]{todonotes}

\icmltitlerunning{InfoDiffusion: Representation Learning Using Information Maximizing Diffusion Models}

\begin{document}

\twocolumn[
\icmltitle{InfoDiffusion: Representation Learning Using \\Information Maximizing Diffusion Models}

\begin{icmlauthorlist}
\icmlauthor{Yingheng Wang}{cu}
\icmlauthor{Yair Schiff}{cu,ct}
\icmlauthor{Aaron Gokaslan}{cu,ct}
\icmlauthor{Weishen Pan}{weill} \\
\icmlauthor{Fei Wang}{weill}
\icmlauthor{Christopher De Sa}{cu}
\icmlauthor{Volodymyr Kuleshov}{cu,ct}
\end{icmlauthorlist}

\icmlaffiliation{cu}{Department of Computer Science, Cornell University, Ithaca, NY, USA}
\icmlaffiliation{ct}{Department of Computer Science, Cornell Tech, New York City, NY, USA}
\icmlaffiliation{weill}{Department of Population Health Sciences, Weill Cornell Medicine, New York City, NY, USA}

\icmlcorrespondingauthor{Yingheng Wang}{yingheng@cs.cornell.edu}

\icmlkeywords{Machine Learning, Generative Modeling, Representation Learning, Diffusion Models, Latent Variable Models}

\vskip 0.3in
]

\printAffiliationsAndNotice{}  %

\begin{abstract}

While diffusion models excel at generating high-quality samples, their latent variables typically lack semantic meaning and are not suitable for representation learning.
Here, we propose InfoDiffusion, an algorithm
that augments diffusion models with low-dimensional latent variables that capture high-level factors of variation in the data. %
InfoDiffusion relies on a learning objective
regularized with the mutual information between observed and hidden variables, which is designed to improve latent space quality
and to prevent the latents from being ignored by expressive diffusion-based decoders.
Empirically, we find that
InfoDiffusion learns
disentangled and human-interpretable latent representations that are competitive with state-of-the-art generative and contrastive methods, while retaining the high sample quality of diffusion models.
Our method enables manipulating the attributes of generated images and has the potential to assist tasks that require exploring a learned latent space to generate quality samples, e.g., generative design.
Code is available on the project page:\par
{\centering\projecturl{https://yingheng-wang.com/infodiffusion-page/}\par}
\end{abstract}
\begin{figure}[ht]
    \centering
    \includegraphics[scale=0.75]{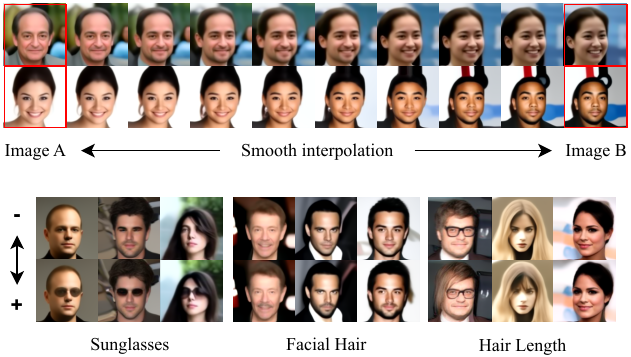}
    \vspace{-0.2cm}
    \caption{
    \textbf{InfoDiffusion produces semantically meaningful latent space for a diffusion model.}
    (\textit{Top}) Smooth latent space. (\textit{Bottom}) Disentangled, human-interpretable factors of variation.
    }
    \label{fig:graph_abs}
\end{figure}

\section{Introduction}\label{sec:intro}
Diffusion models are a family of generative models characterized by high sample quality
\cite{ho2020denoising,dhariwal2021diffusion,rombach2021highresolution}.
These models achieve state-of-the-art performance across a range of generative tasks, including image generation \cite{dhariwal2021diffusion, ramesh2022hierarchical}, audio synthesis \cite{kong2020diffwave}, and molecule design \cite{jing2022torsional, xu2022geodiff}.

However, diffusion models rely on latent variables that typically lack semantic meaning and are not well-suited for the task of representation learning \cite{yang2022diffusion}---the unsupervised discovery of high-level concepts in data (e.g., topics across news articles, facial features in human photos, clusters of related molecules).
This paper seeks to endow diffusion models with a semantically meaningful latent space while retaining their high sample quality.

Specifically, we propose InfoDiffusion, an algorithm
that augments diffusion models with low-dimensional latent variables that capture high-level factors of variation in the data.
InfoDiffusion relies on variational inference and optimizes an objective that is regularized with the mutual information between the low-dimensional latents and the data under the encoder distribution, together with a divergence that matches the aggregate posterior to the prior \citep{zhao2017infovae}.
This prevents expressive diffusion-based  generators from ignoring auxiliary latents and promotes their use for storing semantically meaningful and disentangled information \citep{chen2016infogan}.

\begin{figure}
    \centering
    \includegraphics[width=0.85\linewidth, trim=0 21 27 0, clip]{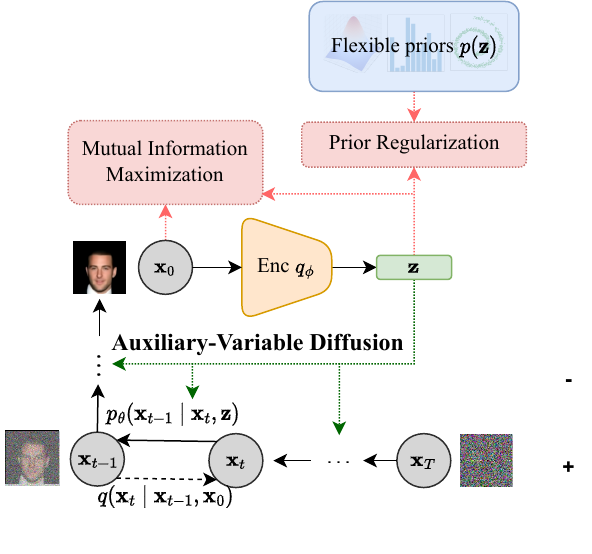}
    \caption{
    Flow chart demonstrating auxiliary-variable diffusion model with mutual information and prior regularization.
    }
    \label{fig:flow_abs}
\end{figure}

The InfoDiffusion algorithm generalizes several existing methods for representation learning
\citep{kingma2013auto,makhzani2015adversarial,higgins2017beta}.
Our method is a principled probabilistic extension of DiffAE \citep{preechakul2022diffusion} that supports custom priors and discrete latents and improves latents via mutual information regularization.
It also extends InfoVAEs \citep{zhao2017infovae} to leverage more flexible diffusion-based decoders.
See \Cref{fig:flow_abs} for an overview of our method.

We evaluate InfoDiffusion on a suite of benchmark datasets and find that it
learns latent representations that are competitive with state-of-the-art generative and contrastive methods \cite{chen2020simple, chen2020improved, caron2021emerging}, while retaining the high sample quality of diffusion models.
InfoDiffusion also finds representations in which individual latent dimensions capture distinct human-interpretable factors of variation; see \Cref{fig:graph_abs} for examples.
\paragraph{Contributions}
In summary, we make the following contributions: (1) we propose a principled probabilistic extension of diffusion models that supports low-dimensional latents; (2) we introduce associated variational learning objectives that are regularized with a mutual information term; (3) we show that these algorithms simultaneously yield high-quality samples and latent representations, achieving competitive performance with state-of-the-art methods on both fronts.

\section{Background}\label{sec:background}
A diffusion model defines a latent variable model $p(\bx_{0:T})$ over data $\bx_0$ sampled from the data distribution, %
as well as latents $\bx_{1:T} := \bx_1, \bx_2, \ldots, \bx_T$ that represent a gradual transformation of $\bx_0$ into  random Gaussian noise $\bx_T$.
The distribution $p$ factorizes as a Markov chain
\begin{equation}
p(\bx_{0:T}) = p(\bx_{T}) \prod_{t=0}^{T-1} p_\theta(\bx_t \mid \bx_{t+1}),
\end{equation}
that maps noise $\bx_T$ into data $\bx_0$ by ``undoing" a noising (or diffusion) process denoted by $q.$
Here we use a learned denoising distribution $p_\theta,$ which we parameterize by a neural network with parameters $\theta$.

The noising process $q$ starts from a clean $\bx_0,$ drawn from the data distribution (denoted by $q(\bx_0)$) and defines a sequence of $T$ variables $\bx_1, ..., \bx_T$ via a Markov chain that factorizes as
\begin{equation}
q(\bx_{1:T} \mid \bx_0) = \prod_{t=1}^T q(\bx_t \mid \bx_{t-1}).
\end{equation}
In this factorization, we define $q(\bx_t \mid \bx_{t-1}) = \mathcal{N}(\bx_t; \sqrt{\alpha_t} \bx_{t-1}, (1-\alpha_t) \mathbf{I})$ as a Gaussian distribution centered around a progressively corrupted version of $\bx_{t-1}$ with a schedule $\alpha_1, \alpha_2, ..., \alpha_T$, where $\mathcal{N}(\cdot\,; \bmu, \bSigma)$ is parameterized by its mean and covariance.
As shown in \citet{ho2020denoising}, the marginal distribution of $q$ can be expressed as
\vspace{0.1cm}
$$
q(\bx_t \mid \bx_0) = \mathcal{N}(\bx_t; \sqrt{\bar \alpha_t} \bx_0, (1-\bar \alpha_t) \mathbf{I}),
$$
\vspace{0.1cm}
where $\bar\alpha_t = \prod_{s=1}^t \alpha_s$ is the cumulative product of the schedule parameters $\alpha_t$.

Normally, $p$ is trained via maximization of an evidence lower bound (ELBO) objective derived using variational inference:
\begin{multline*}
\log p(\bx_{0}) \geq \Exp_{q(\bx_{1} \mid \bx_{0})} [\log p_\theta(\bx_{0} \mid \bx_{1})] \\
- \KL(q(\bx_T \mid \bx_{0}) || p(\bx_T)) \\
- \sum_{t=2}^T \Exp_{q(\bx_{t} \mid \bx_{0})} \left[ \KL(q(\bx_{t-1} \mid \bx_{t}, \bx_0) || p_\theta(\bx_{t-1} \mid \bx_{t})) \right],
\end{multline*}
where $\KL$ denotes the Kullback–Leibler divergence.

\paragraph{Unsupervised Representation Learning}
A core aim of generative modeling is representation learning, the unsupervised extraction of latent concepts from data.
Generative models $p(\bx, \bz)$ typically represent latent concepts using low-dimensional variables $\bz$ that are inferred via posterior inference over $p(\bz \mid \bx)$. %
VAEs exemplify this framework but do not produce state-of-the-art samples.
Conversely, diffusion models produce high-quality samples but lack an interpretable low-dimensional latent space, making them unsuitable for representation learning.

\section{Diffusion Models With Auxiliary Latents}

This paper seeks to endow diffusion models with a semantically meaningful latent space while retaining their high sample quality.
Our strategy is three-fold: (1) in this section, we define a diffusion model family that supports low-dimensional latent variables; (2) in \Cref{sec:4}, we define learning objectives for this model family; (3) in \Cref{sec:5}, we define a regularizer based on mutual information that further encourages the model to learn high-quality latents.

Specifically, we define an auxiliary-variable diffusion model as a probability distribution $p(\bx_{0:T}, \bz)$ that factorizes as:
\begin{equation}
p(\bx_{0:T}, \bz) = p(\bx_{T}) p(\bz) \prod_{t=1}^{T} p_\theta(\bx_{t-1} \mid \bx_{t}, \bz).
\end{equation}
This model implements a reverse diffusion process $p_\theta(\bx_{t-1} \mid \bx_{t}, \bz)$ over $\bx_{0:T}$ conditioned on auxiliary latents $\bz$ distributed according to a prior $p(\bz)$.
The forward (noising) process $q(\bx_{1:T}\mid\bx_0)$ does not depend on $\bz$, because $\bz$ is meant to be a latent representation of the input, not a control variable of diffusion.

\subsection{Auxiliary Latent Variables and Semantic Prior}
The goal of the auxiliary latents $\bz$ is to encode a high-level representation of $\bx_0$.
Unlike $\bx_{1:T}$, the $\bz$ are not constrained to have a particular dimension and can represent a low-dimensional vector of latent factors of variation. They can be continuous, as well as discrete.

The prior $p(\bz)$ ensures that we have a principled probabilistic model and enables the unconditional sampling of $\bx_0$.
The prior can also be used to encode domain knowledge about $\bz$---e.g., if we know that the dataset contains $K$ distinct classes, we may set $p(\bz)$ to be a mixture of $K$ components.
Alternatively, we may set $p(\bz)$ to be a simple distribution from which we can easily sample (e.g., a Gaussian). %

\subsection{Auxiliary-Variable Diffusion Decoder}

The decoder $p_\theta(\bx_{t-1} \mid \bx_{t}, \bz)$ is conditioned on the auxiliary latents $\bz$.
In a trained model, we expect the $\bz$ to be responsible for high-level concepts (e.g., the age or skin color of a person), while the sequence of $\bx_t$ progressively adds lower-level details (e.g., hair texture).

Following previous work \cite{ho2020denoising}, we define the decoder as a Gaussian $p_\theta(\bx_{t-1} \mid \bx_{t}, \bz) = \mathcal{N}(\bx_{t-1}; \bmu_\theta(\bx_t, t, \bz), \sigma_t^2\mathbf{I})$ with fixed variance $\sigma_t^2$ and mean
\begin{align*}
    \bmu_\theta(\bx_{t}, t, \bz) = \frac{1}{\sqrt{\alpha_{t}}} \left( \bx_{t} - \frac{1-\alpha_{t}}{\sqrt{1-\bar \alpha_{t}}} \epsilon_\theta(\bx_{t}, t, \bz) \right)
\end{align*}
with a noise prediction network $\epsilon_\theta(\bx_{t}, t, \bz)$ parameterized by a U-Net \cite{ronneberger2015u}.
We condition this network on $\bz$ using adaptive group normalization layers (AGN), inspired by \citet{dhariwal2021diffusion},
\begin{align*}
    \text{AGN}(\bh, \bz) = (1 + \bs(\bz)) \cdot \text{GroupNorm}(\bh) + \bb(\bz).
\end{align*}
Specifically, in each residual block we first apply AGN with the time embedding and then a second scale-and-shift modulation with the embedding of $\bz$ (without re-normalizing), i.e., $\bh \leftarrow (1+\bs(\bz))\cdot\text{AGN}(\bh, t) + \bb(\bz)$, where the embedding of $\bz$ is produced by a two-layer MLP and is shared by all residual blocks of the U-Net (\Cref{app:arch}).

\paragraph{Continuous-time parameterization} In our implementation, we use the continuous-time limit of this model with the linear noising process $\bx_t = (1-t)\bx_0 + t\bepsilon$, $\bepsilon\sim\mathcal{N}(\mathbf{0},\mathbf{I})$, $t\in[0,1]$ \citep{liu2023flow,lipman2023flow}. This is a Gaussian diffusion with signal-to-noise ratio $(1-t)^2/t^2$; we write $\bx_T$ for its noise endpoint $\bx_{t=1}$. The network predicts the velocity $\bv_\theta(\bx_t, t, \bz) \approx \bepsilon - \bx_0$, from which the noise and data predictions follow as $\epsilon_\theta = \bx_t + (1-t)\bv_\theta$ and $\hat\bx_0 = \bx_t - t\bv_\theta$, so the decoder means above are obtained by substitution. The denoising terms of the ELBO then reduce to a weighted regression loss $\Exp_{t,\bepsilon}[w(t)\,\|\bv_\theta(\bx_t,t,\bz) - (\bepsilon-\bx_0)\|^2]$, where the weighting $w(t)$ is determined by the distribution from which $t$ is sampled. With uniform $t$, the implied weighting is monotone in the noise level, and the objective equals the ELBO under Gaussian data augmentation \citep{kingma2023understanding}. On most datasets, we instead sample $t$ from a logit-normal distribution \citep{esser2024scaling}, which yields a non-monotone reweighting of $\Lc_D$; we treat this choice as a hyperparameter (\Cref{app:exp_details}).

\section{Learning and Inference Algorithms For Auxiliary-Variable Diffusion Models}\label{sec:4}

Next, we introduce learning algorithms for auxiliary-variable models based on variational inference.
We refer to the resulting method as variational auxiliary-variable diffusion.

\subsection{Variational Inference for Auxiliary-Variable Models}

We apply variational inference twice to form a variational lower bound on the marginal log-likelihood of the data (see the full derivation in \Cref{app:prop1}):
\begin{align}\label{eq:denoise}
\log &p(\bx_0) = \log~\Exp_{q_\bz} \Bigg[\frac{p(\bx_0, \bz)}{q_\phi(\bz\mid\bx_0)}\Bigg] \nonumber \\
&\geq\Exp_{q_\bz}\Bigg[\log ~\Exp_{q_\bx} \Bigg[\frac{p(\bx_{0:T}, \bz)}{q_\phi(\bz \mid \bx_0) q(\bx_{1:T} \mid \bx_0)}\Bigg]\Bigg] \nonumber \\
&\geq \Exp_{q_\bx}\Bigg[\Exp_{q_\bz}\Bigg[\log \frac{p(\bx_{0:T}, \bz)}{q_\phi(\bz \mid \bx_0) q(\bx_{1:T} \mid \bx_0)}\Bigg]\Bigg] \nonumber \\
\begin{split}
&=\Exp_{q_{\bx_1}}\left[\Exp_{q_\bz}\left[\log p_{\theta}(\bx_0\mid\bx_1, \bz)\right]\right] -\KL({q_\phi(\bz \mid \bx_0)} || {p(\bz)}) \\
&~~~~ - \KL({q(\bx_T \mid \bx_0)} || {p(\bx_T)}) -\sum_{t=2}^{T} \Exp_{q_{\bx_t}}\left[\Exp_{q_\bz}\left[\KL(q_t || p_t)\right]\right]
\end{split} \nonumber \\
&:= \mathcal{L}_D (\bx_0)
\end{align}
where $\Lc_D(\bx_0)$ denotes the ELBO for a variational auxiliary-variable diffusion model,
$q_t, p_t$ denote the distributions $q(\bx_{t-1} \mid \bx_{t}, \bx_0)$ and $p_\theta(\bx_{t-1} \mid \bx_{t}, \bz)$, respectively, $q_\bz := q_\phi(\bz\mid\bx_0)$ is an approximate variational posterior, $q_\bx := q(\bx_{1:T} \mid \bx_0)$, and $q_{\bx_t} := q(\bx_t \mid \bx_0)$.

We optimize the above objective end-to-end using gradient descent by using the reparameterization trick to backpropagate through samples from $q_\phi(\bz \mid \bx_0)$ \cite{kingma2013auto}.
We use a neural network with parameters $\phi$ to encode the parameters of the approximate posterior distribution of $\bz$.
When the per-sample KL term receives zero weight (the setting $\zeta=1$ of \Cref{sec:5}), the objective does not constrain the posterior variance, and we use a deterministic encoder network $\bz = f_\phi(\bx_0)$, i.e., the limit $q_\phi(\bz\mid\bx_0)\to\delta(\bz - f_\phi(\bx_0))$ (apart from the random input crops described below). For $\zeta<1$, $q_\phi(\bz\mid\bx_0)$ is a Gaussian with diagonal covariance.

\subsection{Inferring Latent Representations}
Once the model is trained, we rely on the approximate posterior $q_\phi(\bz\mid\bx_0)$ to infer $\bz$. In our experiments, we parameterize $q_\phi(\bz\mid\bx_0)$ with the downsampling half of a U-Net followed by a linear head. Before the head, the final feature map is either averaged over space or flattened. Flattening preserves the spatial layout, which matters for aligned data such as faces, digits, and synthetic scenes (\Cref{app:arch}). During training, the encoder receives a random resized crop of $\bx_0$ (covering 70--100\% of the image side), while the decoder denoises the full image. The crop adds randomness to $q_\phi(\bz\mid\bx_0)$; with a deterministic encoder ($\zeta=1$), it is the only source of randomness. At inference time, we encode the full image and use the posterior mean. We do not use crops on 3DShapes, where object scale is a ground-truth factor.

Additionally, we may encode $\bx_0$ into a latent-variable $\bx_T$, which contains information not captured by the auxiliary variable $\bz$---usually details such as texture and high-frequency components.
Our method integrates the probability-flow ODE $\mathrm{d}\bx_t/\mathrm{d}t = \bv_\theta(\bx_t, t, \bz)$ of the learned decoder forward in time, from $t=0$ to $t=1$, conditioned on the inferred $\bz$. This is the continuous-time analogue of running the deterministic DDIM process in reverse \citep{song2021denoising}. Decoding $(\bz, \bx_T)$ with the same ODE in the opposite direction approximately reconstructs $\bx_0$.

\subsection{Discrete Auxiliary-Variable Diffusion}\label{subsec:discrete}

In many settings, latent representations are inherently discrete---e.g., the presence of certain objects in a scene, the choice of topic in a text, etc. Variational auxiliary-variable diffusion supports such discrete variables, e.g., via relaxation methods for deep latent variable models \cite{jang2016categorical} or via quantization of the encoder output \citep{van2017neural,mentzer2024finite}.

In our experiments, we use finite scalar quantization (FSQ) \citep{mentzer2024finite}. The encoder output $\bh = f_\phi(\bx_0)\in\mathbb{R}^{d_z}$ is bounded and rounded to $L$ levels per dimension, $z_i = \mathrm{round}(c\tanh(h_i))/c$ with $c=(L-1)/2$, so that each $z_i$ takes one of $L$ equally spaced values in $[-1,1]$. Gradients are passed through the rounding with the straight-through estimator. This is a deterministic encoder onto a finite set of $L^{d_z}$ codes, the discrete analogue of the deterministic encoder used for $\zeta=1$ (\Cref{sec:4}), and it requires neither a codebook nor auxiliary losses. We use a uniform prior over the codes. For a deterministic encoder, the per-sample KL divergence to this prior is constant, so in the $(\lambda,\zeta)=(0.1,0.9)$ setting no regularization term remains. Relaxation-based variants, in which $\bz$ is replaced during training by a Gumbel-Softmax sample with temperature $\tau$ \citep{jang2016categorical,maddison2017concrete}, did not learn informative codes with our decoder (\Cref{app:discrete}).
See \Cref{app:discrete} for more details.

\subsection{Sampling Methods}\label{subsec:sampling}
At inference time, our model supports multiple sampling procedures.
First, to generate $\bx_0$ unconditionally, we can sample from the original prior $p(\bz),$ as in a VAE, draw $\bx_T\sim\mathcal{N}(\mathbf{0},\mathbf{I})$, and integrate the probability-flow ODE from $t=1$ to $t=0$ with Heun's method \citep{karras2022elucidating} (see \Cref{app:sampling_prior}).
Alternatively, we can utilize a learned prior to potentially improve sample quality (see \Cref{app:sampling_learned} for details on implementing the learned prior used in \Cref{sec:experiments}).
This learned prior is similar to the approach described in DiffAE \cite{preechakul2022diffusion}, where a latent diffusion model is required to enable sampling.
Both procedures can be combined with classifier-free guidance on $\bz$ \citep{ho2022classifier}. To support guidance, we train our generative models ($d_z=256$) with latent dropout: with probability $0.1$, the embedding of $\bz$ is replaced by a learned null embedding $\varnothing$ (the models used to evaluate latents are trained without it). At sampling time, we use the velocity $\bv_\theta(\bx_t,t,\varnothing) + w\,(\bv_\theta(\bx_t,t,\bz) - \bv_\theta(\bx_t,t,\varnothing))$ with guidance weight $w\ge 1$; $w=1$ recovers the conditional decoder.

\section{InfoDiffusion: Regularizing Semantic Latents By Maximizing Mutual Information
}\label{sec:5}

Diffusion models with auxiliary latents face two risks.
First, an expressive decoder $p_\theta(\bx_{t-1}\mid\bx_t, \bz)$ may choose to ignore low-dimensional latents $\bz$ and generate $\bx_{t-1}$ unconditionally \cite{chen2016infogan}.
Second, the aggregate approximate posterior $q_\phi(\bz) = \Exp_{q(\bx_0)}[q_\phi(\bz\mid\bx_0)]$ may fail to match the prior $p(\bz)$ because the prior regularization term is too weak relative to the reconstruction term \cite{zhao2017infovae}. This degrades the quality of ancestral sampling as well as that of latent representations.

\subsection{Regularizing Auxiliary-Variable Diffusion}

We propose dealing with the issues of ignored latents and degenerate posteriors by using two regularization terms---a mutual information term and a prior regularizer. We refer to the resulting algorithm as InfoDiffusion.

\paragraph{Mutual Information Regularization}

To prevent the diffusion model from ignoring the latents $\bz$, we augment the learning objective from \Cref{eq:denoise} with a mutual information term \cite{chen2016infogan,zhao2017infovae} between $\bx_0$ and $\bz$ under $q_\phi(\bx_0, \bz),$
the joint distribution over observed data $\bx_0$ and latent variables $\bz$.
Formally, we define the mutual information regularizer as
$$
\MI_{\bx_0, \bz} = \mathbb{E}_{q_\phi(\bx_0, \bz)}\left[ \log \frac{q_\phi(\bx_0, \bz)}{q(\bx_0)q_\phi(\bz)} \right]
$$
where $q_\phi(\bz)$ is the marginal approximate posterior distribution---defined as the marginal of the product $q_\phi(\bz\mid\bx_0)q(\bx_0).$
Intuitively, maximizing mutual information encourages the encoder to produce latents $\bz$ that are predictive of $\bx_0$ (and vice versa), which counteracts the per-sample KL term of $\Lc_D$ that otherwise penalizes $\MI_{\bx_0,\bz}$ (see \Cref{eq:25} in \Cref{app:prop1}).

\paragraph{Prior Regularization}

To prevent the model from learning a degenerate approximate posterior, we regularize the encoded samples $\bz$ to look like the prior $p$. Formally, we define the prior regularizer as
$$
\Rc = \text{D}(q_\phi(\bz)||p(\bz)),
$$
where $\text{D}$ is any strict divergence.

\subsection{A Tractable Objective for InfoDiffusion}
We train InfoDiffusion by maximizing a regularized ELBO objective of the form
\begin{equation}\label{eq:infodiff}
\mathbb{E}_{q(\bx_0)}[\Lc_D(\bx_0)] + \zeta \cdot \MI_{\bx_0, \bz} - \beta \cdot \Rc,
\end{equation}
where $\Lc_D(\bx_0)$ is from \Cref{eq:denoise}, and $\zeta, \beta \in \mathbb{R}$ are scalars controlling the strength of the regularizers.
Note that $\mathbb{E}_{q(\bx_0)}[\Lc_D(\bx_0)]$ already contains $-\KL(q_\phi(\bz)\|p(\bz)) - \MI_{\bx_0,\bz}$ (\Cref{eq:25}); with $\Rc = \KL(q_\phi(\bz)\|p(\bz))$, the total weight on the marginal divergence is therefore $\beta + 1$, and $\beta$ may be negative.

However, both the mutual information and the prior regularizer are intractable.
Following \citet{zhao2017infovae}, we rewrite the above learning objective into an equivalent tractable form, as described in \Cref{prop:reg_infodiff} (see \Cref{app:prop1} for the full derivation).
Defining $\lambda := \beta + 1$ (the total weight on $\KL(q_\phi(\bz)\|p(\bz))$), we have

\begin{proposition}\label{prop:reg_infodiff}
    The regularized InfoDiffusion objective, \Cref{eq:infodiff}, can be rewritten as
    \begin{align}\label{eq:obj}
        &\Lc_I = \Exp_{q(\bx_0, \bx_1)}\left[\Exp_{q_\bz}\left[\log p_{\theta}(\bx_0|\bx_1, \bz)\right]\right] \nonumber \\
        &\enskip- \mathbb{E}_{q(\bx_0)}[\KL({q(\bx_T | \bx_0)} || {p(\bx_T)})] \nonumber  \\
        &\enskip- \sum_{t=2}^{T} \Exp_{q(\bx_0, \bx_{t})}\left[\Exp_{q_\bz} \left[\KL({q(\bx_{t-1}|\bx_t, \bx_0)} ||{p_{\theta}(\bx_{t-1}|\bx_t, \bz)})\right]\right] \nonumber  \\
        &\enskip- (1-\zeta) \mathbb{E}_{q(\bx_0)} [\KL(q_\phi(\bz\mid\bx_0)||p(\bz))] \nonumber  \\
        &\enskip- (\lambda + \zeta-1) \KL(q_\phi(\bz) || p(\bz))
    \end{align}
\end{proposition}
We now state that $\KL(q_\phi(\bz) || p(\bz))$ from \Cref{eq:obj} can be replaced with any strict divergence $\D(q_\phi(\bz) || p(\bz))$ without modifying the original objective in \Cref{prop:divergence} (see \Cref{app:prop2} for the full derivation).

\begin{proposition}\label{prop:divergence}
The term $\KL(q_\phi(\bz)|| p(\bz))$ in \Cref{prop:reg_infodiff} can be replaced with any strict divergence term $\text{D}(q_\phi(\bz)|| p(\bz))$ and meanwhile the InfoDiffusion objective $\Lc_I$ is guaranteed to be globally optimized for any fixed value ${I}_0$ of $\MI_{\bx_0, \bz}$ when input space $\mathcal{X}_0$ and feature space $\mathcal{Z}$ are continuous spaces, $\zeta \leq 1$, $\lambda + \zeta - 1 \geq 0$, if $p_{\theta}(\bx_{t-1}|\bx_t, \bz) = q_\phi(\bx_{t-1}|\bx_t, \bz)$ and $q_\phi(\bz) = p(\bz)$, where $q_\phi(\bx_{t-1}|\bx_t, \bz)$ is the reverse conditional of the joint $q(\bx_0)q_\phi(\bz\mid\bx_0)q(\bx_{1:T}\mid\bx_0)$; this holds exactly in the limit $\bar\alpha_T\to 0$, and up to a term bounded by $\MI(\bx_T;\bx_0)$ otherwise.
\end{proposition}
Thus, there are a range of divergences that can be compatible with our framework.
In our experiments, we consider the maximum mean discrepancy (MMD) \cite{gretton2012kernel}, defined as:
\begin{align*}
    \mathrm{MMD}(q_\phi(\bz) || p(\bz)) &= \mathbb{E}_{\bz,\bz'\sim q_\phi(\bz)} [k(\bz, \bz')] \\
    &+ \mathbb{E}_{\bz,\bz'\sim p(\bz)}[k(\bz, \bz')] \\
    &- 2\mathbb{E}_{\bz\sim q_\phi(\bz),\bz'\sim p(\bz)}[k(\bz, \bz')]
\end{align*}
where $k$ is a positive definite kernel.
In order to optimize $\text{MMD}(q_\phi(\bz) || p(\bz))$, we use sample-based optimization methods for implicit models.
Specifically we estimate expectations over $q_\phi(\bz)$ by taking empirical averages over the encodings of samples $\{\bx_0^{(i)}\}_{i=1}^N \sim q(\bx_0)$ in each mini-batch, and expectations over $p(\bz)$ with an equal number of prior samples. Following \citet{tolstikhin2018wasserstein}, we use a sum of inverse multiquadratic kernels, $k(\bz,\bz') = \sum_{s} C_s/(C_s + \|\bz-\bz'\|^2)$ with $C_s = 2 d_z s$ for $s\in\{0.1, 0.2, 0.5, 1, 2, 5, 10\}$, where $d_z$ is the latent dimension, and the unbiased (U-statistic) estimator. The heavy tails of this kernel keep it sensitive to both the location and the scale of $q_\phi(\bz)$ across a wide range of distances.

\paragraph{Loss weights} We normalize the objective per data dimension: the denoising loss is averaged over the $D_x$ pixel dimensions, and the per-sample KL term is divided by $D_x$, so that $(1-\zeta)$ is its weight relative to the denoising loss, exactly as in \Cref{eq:obj}. Since the MMD is measured on a different scale than the KL divergence, we use $(\lambda+\zeta-1)$ directly as the weight of the MMD estimate; this weight is therefore not directly comparable to $(1-\zeta)$.

\subsection{Comparing InfoDiffusion to Existing Models}

The InfoDiffusion algorithm generalizes several existing methods in the literature.
When the decoder performs one step of diffusion ($T=1$), we recover a model that is equivalent to the InfoVAE model \cite{zhao2017infovae}, up to choices of the decoder architecture.
When we additionally choose $\lambda = 1-\zeta$, so that the marginal divergence term vanishes, we recover the $\beta$-VAE model \cite{higgins2017beta} with $\beta_{\text{VAE}} = 1-\zeta$ (the standard VAE corresponds to $\zeta=0$, $\lambda=1$; note that $\beta_{\text{VAE}}>1$ corresponds to $\zeta<0$, i.e., a penalty on mutual information).
When $T=1$, $\zeta=1$, and $\D$ is the Jensen-Shannon divergence (estimated adversarially), we recover adversarial auto-encoders (AAEs) \cite{makhzani2015adversarial}; with $T=1$, $\zeta=1$ and $\D=\mathrm{MMD}$, we recover the MMD-based Wasserstein auto-encoder \citep{tolstikhin2018wasserstein}.
Our InfoDiffusion method can be seen as an extension of $\beta$-VAE, InfoVAE, and AAE to diffusion decoders, similar to how denoising diffusion probabilistic models (DDPM; \citet{ho2020denoising}) extend VAEs.
Finally, when $\zeta=1$ and $\lambda=0$, both regularizers vanish and, with a deterministic encoder, we recover the DiffAE model \cite{preechakul2022diffusion}. The configuration $(\lambda,\zeta)=(0.1,0.9)$ used in our main experiments satisfies $\lambda=1-\zeta$: the marginal divergence term vanishes, and the objective is that of a $\beta$-VAE with $\beta_{\text{VAE}}=0.1$ and a diffusion decoder. With $\zeta=1$ and $\lambda>0$, InfoDiffusion instead corresponds to DiffAE augmented with an MMD prior-matching term on the aggregate posterior.
We further discuss how our method relates to these prior works in \Cref{sec:related}.
In \Cref{tab:comp}, we detail this comparison to special cases.
\begin{table}[t]
    \caption{Comparison of InfoDiffusion model to other auto-encoder (\textit{top}) and diffusion (\textit{bottom}) frameworks in terms of enabling semantic latents, discrete latents, custom priors, mutual information maximization (Max MI), and high-quality sample generation (HQ samples). $\beta$-VAE with $\beta_{\text{VAE}}>1$ penalizes rather than maximizes mutual information; AAEs support categorical latents through adversarial matching of a categorical prior \citep{makhzani2015adversarial}.}
\newcommand{\greencheck}{{\bf \color{OliveGreen}\checkmark}}

    \newcommand*\colourcheck[1]{%
      \expandafter\newcommand\csname #1check\endcsname{\textcolor{#1}{\ding{52}}}%
    }
    \definecolor{bloodred}{HTML}{B00000}
    \definecolor{cautionyellow}{HTML}{EED202}
    \newcommand*\colourxmark[1]{%
      \expandafter\newcommand\csname #1xmark\endcsname{\textcolor{#1}{\ding{54}}}%
    }
    \newcommand*\colourcheckodd[1]{%
      \expandafter\newcommand\csname #1checkodd\endcsname{\textcolor{#1}{\ding{51}}}%
    }
    \colourcheckodd{cautionyellow}
    \colourcheck{cautionyellow}
    \colourcheck{OliveGreen}
    \colourxmark{bloodred}

    \newcommand{\ourxmark}{\bloodredxmark}%
    \newcommand{\ourcheckmark}{\OliveGreencheck}

    \label{tab:comp}
    \begin{center}
    \begin{small}
        \setlength{\tabcolsep}{4.5pt}
        \begin{tabular}{lcccccc}
    \toprule
        & \makecell{Semantic\\latents} & \makecell{Discrete\\latents} & \makecell{Custom\\prior} & \makecell{Max\\MI} & \makecell{HQ\\samples} \\
        \midrule
        AE & \ourxmark & \ourxmark  & \ourxmark & \ourxmark & \ourxmark \\
         VAE & \ourcheckmark & \ourcheckmark  & \ourcheckmark & \ourxmark & \ourxmark \\
         $\beta$-VAE & \ourcheckmark & \ourcheckmark & \ourcheckmark & \ourxmark & \ourxmark \\
         AAE & \ourcheckmark & \ourcheckmark  & \ourcheckmark & \ourcheckmark & \ourxmark \\
         InfoVAE & \ourcheckmark & \ourcheckmark  & \ourcheckmark & \ourcheckmark & \ourxmark \\
         \midrule
         DDPM & \ourxmark & \ourxmark  & \ourxmark & \ourxmark & \ourcheckmark \\
         DiffAE & \ourcheckmark & \ourxmark  & \ourxmark & \ourxmark & \ourcheckmark \\
         InfoDiff & \ourcheckmark & \ourcheckmark  & \ourcheckmark & \ourcheckmark & \ourcheckmark \\
         \bottomrule
    \end{tabular}
    \end{small}
    \end{center}

\end{table}

\section{Experiments} \label{sec:experiments}
In this section, we evaluate our proposed method by comparing it to several baselines, using metrics that span generation quality, utility of latent space representations, and disentanglement.
The baselines we compare against are: a vanilla auto-encoder (AE) \cite{lecun1987phd}, a VAE \cite{kingma2013auto, higgins2017beta}, an InfoVAE \cite{zhao2017infovae}, and a DiffAE \cite{preechakul2022diffusion}.

We measure performance on the following datasets: FashionMNIST \cite{xiao2017fashion}, CIFAR10 \cite{krizhevsky2009learning}, FFHQ \cite{karras2019style}, CelebA \cite{liu2015faceattributes}, and 3DShapes \cite{3dshapes18}.
See \Cref{app:exp_details} for complete hyperparameter and computational resource details, by dataset.
Unless stated otherwise, InfoDiffusion uses $\lambda=0.1$ and $\zeta=0.9$, the best setting in our evaluation of the regularization coefficients (\Cref{subsec:choices}). We evaluate latents with $d_z=32$ models ($d_z=10$ on 3DShapes) trained without latent dropout, and sample quality with $d_z=256$ models trained with latent dropout (\Cref{subsec:sampling}). InfoDiffusion results are averaged over three training seeds. We re-train DiffAE on every dataset with its official implementation, and the AE, VAE, $\beta$-VAE, and InfoVAE on FFHQ with the released InfoDiffusion code; the remaining baseline results come from earlier runs with a separate training pipeline (\Cref{app:exp_details}). As discussed in \Cref{subsec:sampling}, for InfoDiffusion, we experiment with generating images using either $\bz$ drawn from the prior or drawn from a learned latent distribution (denoted as ``w/Learned Latent'' in \Cref{tab:class} and \Cref{tab:disentangle}, see \Cref{app:sampling_learned} for details), without guidance ($w=1$) or with guidance on $\bz$ ($w=2$).

\subsection{Exploring Latent Representations}
We start by exploring three qualitative desirable features of learned representations: (1) their ability to capture high level semantic content, (2) smooth interpolation in latent space translating to smooth changes in generated output, and (3) their utility in downstream tasks.
\begin{figure}[ht]
    \centering

    \setlength{\tabcolsep}{0pt} %
    \setlength{\itemwidth}{0.15\linewidth}
    \renewcommand{\arraystretch}{0}
    \newlength{\rowheight} %

    \begin{tabular}{cccccc}
        && \multicolumn{4}{c}{\small{~~~~\leftarrowfill~~~~~~~~Fixed $\bz,$ varying $\bx_T$~~~~~~~~\rightarrowfill}}\\
        \settoheight{\rowheight}{\includegraphics[width=\itemwidth, cfbox=red 1pt 0pt]{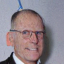}} %
        \raisebox{0.5\dimexpr\rowheight-\height}{Image 1}~~~ & \includegraphics[width=\itemwidth, cfbox=red 1pt 0pt]{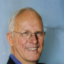} &
        \includegraphics[width=\itemwidth]{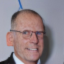} &
        \includegraphics[width=\itemwidth]{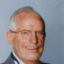} &
        \includegraphics[width=\itemwidth]{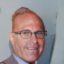} &
        \includegraphics[width=\itemwidth]{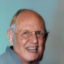}
        \\
        \settoheight{\rowheight}{\includegraphics[width=\itemwidth, cfbox=red 1pt 0pt]{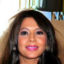}} %
        \raisebox{0.5\dimexpr\rowheight-\height}{Image 2}~~~ & \includegraphics[width=\itemwidth, cfbox=red 1pt 0pt]{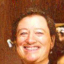} &
        \includegraphics[width=\itemwidth]{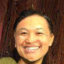} &
        \includegraphics[width=\itemwidth]{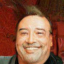} &
        \includegraphics[width=\itemwidth]{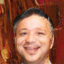} &
        \includegraphics[width=\itemwidth]{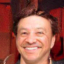}
        \\
        \settoheight{\rowheight}{\includegraphics[width=\itemwidth, cfbox=red 1pt 0pt]{qualitative-figures/a_fixed1/sample-000000.png}} %
        \raisebox{0.5\dimexpr\rowheight-\height}{Image 3}~~~ & \includegraphics[width=\itemwidth, cfbox=red 1pt 0pt]{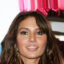} &
        \includegraphics[width=\itemwidth]{qualitative-figures/a_fixed1/sample-000000.png} &
        \includegraphics[width=\itemwidth]{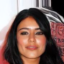} &
        \includegraphics[width=\itemwidth]{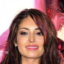} &
        \includegraphics[width=\itemwidth]{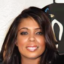}
        \\
    \end{tabular}
    \caption{$\bz$ captures high-level semantic detail. Varying $\bx_T \sim$ $\mathcal{N}(\mathbf{0}, \mathbf{I})$ (across the columns in each row) changes lower level detail in the image. Red box indicates original image.
    }
    \label{fig:semantics}
\end{figure}

\begin{figure*}[ht]
    \centering
\setlength{\tabcolsep}{0pt} %
\setlength{\itemwidth}{0.07\linewidth}
\renewcommand{\arraystretch}{0}

\begin{tabular}{ccccccccccc}

\includegraphics[width=\itemwidth]{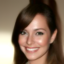} &
\includegraphics[width=\itemwidth]{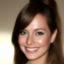} &
\includegraphics[width=\itemwidth]{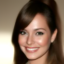} &
\includegraphics[width=\itemwidth]{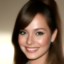} &
\includegraphics[width=\itemwidth]{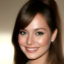} &
\includegraphics[width=\itemwidth]{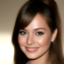} &
\includegraphics[width=\itemwidth]{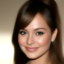} &
\includegraphics[width=\itemwidth]{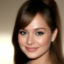} &
\includegraphics[width=\itemwidth]{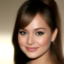} &
\includegraphics[width=\itemwidth]{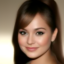} &
\includegraphics[width=\itemwidth]{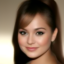}
\\
\vspace{0.1em}\\
\includegraphics[width=\itemwidth]{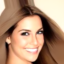} &
\includegraphics[width=\itemwidth]{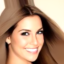} &
\includegraphics[width=\itemwidth]{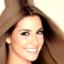} &
\includegraphics[width=\itemwidth]{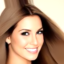} &
\includegraphics[width=\itemwidth]{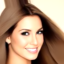} &
\includegraphics[width=\itemwidth]{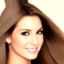} &
\includegraphics[width=\itemwidth]{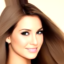} &
\includegraphics[width=\itemwidth]{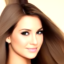} &
\includegraphics[width=\itemwidth]{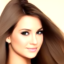} &
\includegraphics[width=\itemwidth]{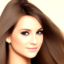} &
\includegraphics[width=\itemwidth]{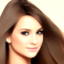}
\\
\vspace{0.3em}\\
\multicolumn{11}{c}{\textbf{+}~~\small{\leftarrowfill~~~~~~~~ Smile~~~~~~~~~\rightarrowfill}~~\textbf{\textendash}}\\

\end{tabular}
\caption{Finding disentangled dimensions in InfoDiffusion's auxiliary latent variable $\bz.$
Images are produced through a linear traversal along a particular dimension, spanning values from -1.5 to 1.5.
}
    \label{fig: disentangle}
\end{figure*}

\begin{figure*}[ht]
    \centering
\setlength{\tabcolsep}{0pt} %
\setlength{\itemwidth}{0.07\linewidth}
\renewcommand{\arraystretch}{0}

\newlength{\arrowlength}
\setlength{\arrowlength}{\textwidth-7\itemwidth}

\begin{tabular}{cccccccccc}
Image A
& \multicolumn{8}{c}{$\xleftrightarrow{\makebox[\arrowlength]{}}$} &
Image B \\
\vspace{0.3em} \\
\includegraphics[width=\itemwidth, cfbox=red 1pt 0pt]{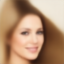} &
\includegraphics[width=\itemwidth]{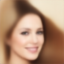} &
\includegraphics[width=\itemwidth]{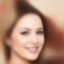} &
\includegraphics[width=\itemwidth]{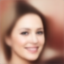} &
\includegraphics[width=\itemwidth]{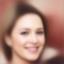} &
\includegraphics[width=\itemwidth]{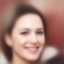} &
\includegraphics[width=\itemwidth]{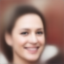} &
\includegraphics[width=\itemwidth]{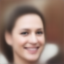} &
\includegraphics[width=\itemwidth]{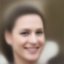} &
\includegraphics[width=\itemwidth, cfbox=red 1pt 0pt]{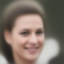}
\\
\vspace{0.3em}\\
\multicolumn{10}{c}{\small{(a) VAE}}\\
\vspace{0.3em}\\
\includegraphics[width=\itemwidth, cfbox=red 1pt 0pt]{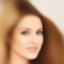} &
\includegraphics[width=\itemwidth]{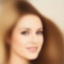} &
\includegraphics[width=\itemwidth]{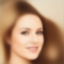} &
\includegraphics[width=\itemwidth]{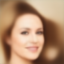} &
\includegraphics[width=\itemwidth]{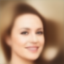} &
\includegraphics[width=\itemwidth]{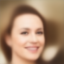} &
\includegraphics[width=\itemwidth]{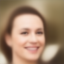} &
\includegraphics[width=\itemwidth]{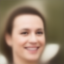} &
\includegraphics[width=\itemwidth]{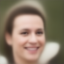} &
\includegraphics[width=\itemwidth, cfbox=red 1pt 0pt]{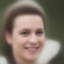}
\\
\vspace{0.3em}\\
\multicolumn{10}{c}{\small{(b) InfoVAE}}\\
\vspace{0.3em}\\
\includegraphics[width=\itemwidth, cfbox=red 1pt 0pt]{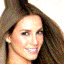} &
\includegraphics[width=\itemwidth]{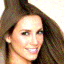} &
\includegraphics[width=\itemwidth]{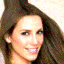} &
\includegraphics[width=\itemwidth]{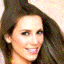} &
\includegraphics[width=\itemwidth]{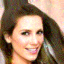} &
\includegraphics[width=\itemwidth]{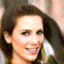} &
\includegraphics[width=\itemwidth]{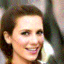} &
\includegraphics[width=\itemwidth]{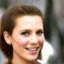} &
\includegraphics[width=\itemwidth]{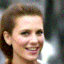} &
\includegraphics[width=\itemwidth, cfbox=red 1pt 0pt]{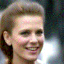}
\\
\vspace{0.3em}\\
\multicolumn{10}{c}{\small{(c) DiffAE}}\\
\vspace{0.3em}\\
\includegraphics[width=\itemwidth, cfbox=red 1pt 0pt]{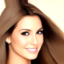} &
\includegraphics[width=\itemwidth]{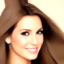} &
\includegraphics[width=\itemwidth]{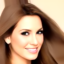} &
\includegraphics[width=\itemwidth]{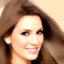} &
\includegraphics[width=\itemwidth]{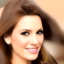} &
\includegraphics[width=\itemwidth]{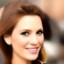} &
\includegraphics[width=\itemwidth]{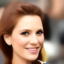} &
\includegraphics[width=\itemwidth]{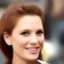} &
\includegraphics[width=\itemwidth]{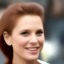} &
\includegraphics[width=\itemwidth, cfbox=red 1pt 0pt]{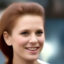}
\\
\vspace{0.3em}\\
\multicolumn{10}{c}{\small{(d) InfoDiffusion}}\\
\end{tabular}
\caption{Latent space interpolation.
InfoDiffusion has a smooth latent space and maintains high image generation quality.
Reconstructions of the two original images are at the left and right ends of each row and are marked by red boxes.}
\label{fig: interpolate}
\end{figure*}

\paragraph{Auxiliary Variables Capture Semantic Information}
In \Cref{fig:semantics}, we demonstrate that our model is able to encode high-level semantic information in the auxiliary variable.
For a fixed $\bz$ and varying $\bx_T$, we find that decoded images change in their low-level features, e.g., background, hair style.

\paragraph{Latent Space Interpolation}
We begin with two images $\bx_0^{(i)}, \bx_0^{(j)}$ and retrieve their corresponding noise and auxiliary latent encodings $(\bz^{(i)}, \bx_T^{(i)}), (\bz^{(j)}, \bx_T^{(j)}).$
Then, for 10 fixed steps $l \in [0, 1],$ we generate images from the latent representations $(\bz^l, \bx_T^l)$ where $\bz^l = \cos(l\pi/2)\bz^{(i)} + \sin(l\pi/2)\bz^{(j)}$ is a trigonometric (quarter-circle) interpolation between the auxiliary latents and $\bx_T^l = [\sin((1-l)\psi)\bx_T^{(i)} + \sin(l\psi)\bx_T^{(j)}]/\sin\psi$ is the spherical interpolation between the noise tensors of the two images, with $\psi$ denoting the angle between $\bx_T^{(i)}$ and $\bx_T^{(j)}.$
In \Cref{fig: interpolate}, we see that our model is able to combine the smooth interpolation of variational methods with the high sample quality of diffusion models.

\paragraph{Latent Variables Discover and Predict Class Labels}
In addition to the qualitative inspection of our latent space, we run downstream classification tasks on $\bz$ to measure its utility, which we report in \Cref{tab:class} and \Cref{tab:disentangle} as ``Latent Qual.''
Specifically, we train a logistic regression classifier on the standardized auxiliary latent encodings of images to predict labels and report the accuracy/AUROC (or average accuracy/AUROC if multiple annotations are predicted) on held-out data.
We use 5-fold cross-validation over the encodings of training images (for CelebA, the first 10{,}000 training images): in each fold, we fit the classifier on 80\% of the encodings and evaluate it on the remaining 20\%, and we report the mean over folds. For InfoDiffusion, we report the mean $\pm$ one standard deviation of this quantity over three training seeds, for the baselines, $\pm$ is the standard deviation across folds. We use third-party labels for age, gender, and eyeglasses for FFHQ (\Cref{app:exp_details}).
We also compute FID (clean-FID \citep{parmar2022aliased}, with statistics of the training set) from 10{,}000 samples. For InfoDiffusion, $\pm$ is the standard deviation over three training seeds, and for the baselines, over five sample sets of a single model (three for the re-trained baselines).

On FashionMNIST, CIFAR10, and FFHQ, the compact latent representations from our models are the most informative of labels.
In addition to the utility of the latent space, we generate high-quality images: with latents drawn from the learned latent prior and guidance on $\bz$, InfoDiffusion attains the lowest FID on CIFAR10 and FFHQ and the second-lowest on FashionMNIST.

\begin{table*}[ht]
\setlength{\tabcolsep}{2pt}
    \caption{Latent quality, as measured by classification accuracies for logistic regression classifiers trained on the auxiliary latent vector $\bz,$ and FID.
    We report mean $\pm$ one standard deviation.
    Darkly shaded cells indicate the best while lightly shaded cells indicate the second best. See \Cref{app:tab:class} for the performance of varying hyperparameters. InfoDiffusion rows without ``w/Learned Latent'' are generated with $\bz\sim p(\bz)$ (\Cref{app:sampling_prior}). Latent quality and FID are measured with $d_z=32$ and $d_z=256$ models, respectively (\Cref{tab:exp_details}). For InfoDiffusion, $\pm$ is the standard deviation over three training seeds, and $w$ is the guidance weight on $\bz$ (\Cref{subsec:sampling}). DiffAE and the FFHQ entries of the other baselines are re-trained; the remaining baseline entries come from earlier runs (\Cref{app:exp_details}).
    }
    \label{tab:class}
    \begin{center}
    \begin{small}
    \begin{sc}
    \begin{tabular}{l @{\hskip 1.5em} cc @{\hskip 1.5em} cc @{\hskip 1.5em} cc}
    \toprule
    &\multicolumn{2}{c}{FashionMNIST}&\multicolumn{2}{c}{CIFAR10}&\multicolumn{2}{c}{FFHQ}\\
    \midrule
    & \makecell{Latent \\Qual. $\uparrow$} & FID $\downarrow$ & \makecell{Latent \\Qual. $\uparrow$} & FID $\downarrow$ &  \makecell{Latent \\Qual. $\uparrow$} & FID $\downarrow$\\
    \midrule
    AE & $ 0.819 \pm 0.003$ & $62.9 \pm 2.1$ & $0.336 \pm 0.005$ & $169. 4 \pm 2.4$ & $0.532 \pm 0.002$ & $316.7 \pm 0.4$\\
    VAE & $0.796 \pm 0.002$ & $63.4 \pm 1.6$ & $ 0.342 \pm 0.004$ & $177.2 \pm 3.2$ & $0.488 \pm 0.002$ & $307.7 \pm 0.1$ \\
    beta-VAE & $ 0.779 \pm 0.004$ & $66.9 \pm 1.8$ & $0.253 \pm 0.003$ & $ 183.3 \pm 3.1$ & $0.510 \pm 0.001$ & $222.6 \pm 0.2$\\
    InfoVAE & $0.807 \pm 0.003$ & $55.0 \pm 1.7$ & $ 0.357 \pm 0.005$ & $ 160.7 \pm 2.5$ & $0.545 \pm 0.001$ & $195.2 \pm 0.5$\\
    DiffAE & \secondplace$0.822 \pm 0.006$ & \firstplace$5.6 \pm 0.1$ & \secondplace$0.442 \pm 0.004$ & $34.1 \pm 0.1$ & \secondplace$0.565 \pm 0.001$ & \secondplace$12.1 \pm 0.1$\\
    \midrule
    InfoDiffusion ($\lambda=0.1,\zeta=0.9$) & \firstplace$0.856 \pm 0.004$ & $22.5 \pm 0.6$ & \firstplace$0.461 \pm 0.003$ & $36.2 \pm 1.1$ & \firstplace$0.628 \pm 0.002$ & $19.8 \pm 0.4$\\
    ~~ w/ guidance ($w=2$) & & $21.9 \pm 0.7$ & & $33.7 \pm 1.0$ & & $15.7 \pm 0.7$\\
    ~~ w/Learned Latent & & $8.1 \pm 0.4$ & & \secondplace$25.7 \pm 0.4$ & & $13.1 \pm 0.5$\\
    ~~ w/Learned Latent, $w=2$ & & \secondplace$6.8 \pm 0.7$ & & \firstplace$21.3 \pm 0.7$ & & \firstplace$9.8 \pm 0.3$\\
    \bottomrule
    \end{tabular}
    \end{sc}
    \end{small}
    \end{center}
\end{table*}

\subsection{Disentanglement}

\subsubsection{Finding disentangled dimensions}
We find that the latent representations learned by InfoDiffusion contain disentangled components.
For example, in \Cref{fig:graph_abs}, we see several examples of disentangled factors.
In \Cref{fig: disentangle}, we demonstrate this in more detail, traversing a specific dimension of $\bz$ that controls smiling from values of -1.5 to 1.5.

\subsubsection{Disentanglement Metrics}
\paragraph{DCI Score} For the 3DShapes dataset, we use the Disentanglement term of the DCI scores proposed in \citet{eastwood2018framework}.
This disentanglement metric is calculated as follows:
for each attribute, a model is trained to predict it using the auxiliary latent vector $\bz$.
The model must also provide the importance of each dimension of $\bz$ in predicting each attribute.
Relative importance weights are converted to probabilities that dimension $i$ of $\bz$ is important for predicting a given label.
The disentanglement score for each dimension of $\bz$ is calculated as 1 minus the entropy of the relative importance probabilities.
If a dimension is important for predicting only a single attribute, the score will be 1.
If a dimension is equally important for predicting all attributes, the disentanglement score will be 0.
The disentanglement scores are then averaged, with weights determined by the relative importance of each dimension across $\bz$, to get the DCI disentanglement score.
In \Cref{tab:disentangle}, we see that for the 3DShapes dataset, InfoDiffusion attains the highest DCI disentanglement score ($0.296$), although the $\beta$-VAE and VAE ($0.281$ and $0.276$) are within one standard deviation across seeds; DCI is sensitive to the regularization coefficients (\Cref{app:coeffs}).

\paragraph{TAD} For the CelebA dataset, we quantify disentanglement using TAD \cite{yeats2022nashae}, which is a disentanglement metric specifically proposed for this dataset that accounts for the presence of correlated and imbalanced attributes.
First, we quantify attribute correlation by calculating the proportion of entropy reduction of each attribute given any other single attribute.
Any attribute with an entropy reduction greater than 0.2 is removed.
For each remaining attribute, we calculate AUROC score of each dimension of the auxiliary latent vector $\bz$ in detecting that attribute.
If an attribute can be detected by at least one dimension of $\bz$, i.e., AUROC $\geq 0.75$, it is considered to be ``captured.''
The TAD score is the summation of the differences of the AUROC between the two most predictive latent dimensions for all captured attributes.
In \Cref{tab:disentangle}, we again see that InfoDiffusion has the best disentanglement performance with more captured attributes ($5.7$ on average) and higher TAD scores ($0.439$).
We additionally note that the InfoDiffusion model balances disentanglement with high-quality generation and good latent space quality.

For calculating DCI on 3DShapes, we follow previous work \cite{locatello2019challenging} and treat the attributes as discrete variables, using a gradient boosting classifier implemented by \texttt{scikit-learn} \cite{scikit-learn} as our predictor.
For disentanglement metric calculation, we fit the DCI predictors on the encodings of 10{,}000 training images and compute importances and scores on 5{,}000 test images. TAD, which does not require a fitted predictor, is computed from per-dimension AUROCs on each of the five training folds used above and averaged over folds.
We report mean metrics $\pm$ one standard deviation over three training seeds for InfoDiffusion.

\begin{table*}[ht]
\setlength{\tabcolsep}{2pt} %
    \caption{Disentanglement and latent quality metrics and FID.
    For 3DShapes, we check the image quality manually and label the models which generate high-quality images with check marks (`Image Qual.').
    The visualization of the samples is shown in \Cref{fig:3dshapes_visual} in \Cref{app:3dshapes}.
    For CelebA, `Attrs.' counts the number of ``captured'' attributes when calculating the TAD score.
    `Latent Quality' is measured as AUROC scores averaged across attributes for logistic regression classifiers trained on the auxiliary latent vector $\bz$.
    We report means $\pm$ one standard deviation for quantitative metrics.
    Darkly shaded cells indicate the best while lightly shaded cells indicate the second best. For InfoDiffusion ($\lambda=0.1$, $\zeta=0.9$), $\pm$ is the standard deviation over three training seeds; FID is measured with $d_z=256$ models, $w$ is the guidance weight on $\bz$, and image quality on 3DShapes is judged from 16 samples drawn with $\bz\sim p(\bz)$ (first row) or from the learned latent prior (\Cref{app:exp_details}). DiffAE is re-trained; the other baseline entries come from earlier runs (\Cref{app:exp_details}).}
    \newcommand{\greencheck}{{\bf \color{OliveGreen}\checkmark}}

    \newcommand*\colourcheck[1]{%
      \expandafter\newcommand\csname #1check\endcsname{\textcolor{#1}{\ding{52}}}%
    }
    \definecolor{bloodred}{HTML}{B00000}
    \definecolor{cautionyellow}{HTML}{EED202}
    \newcommand*\colourxmark[1]{%
      \expandafter\newcommand\csname #1xmark\endcsname{\textcolor{#1}{\ding{54}}}%
    }
    \newcommand*\colourcheckodd[1]{%
      \expandafter\newcommand\csname #1checkodd\endcsname{\textcolor{#1}{\ding{51}}}%
    }
    \colourcheckodd{cautionyellow}
    \colourcheck{cautionyellow}
    \colourcheck{OliveGreen}
    \colourxmark{bloodred}

    \newcommand{\ourxmark}{\bloodredxmark}%
    \newcommand{\ourcheckmark}{\OliveGreencheck}
    \label{tab:disentangle}
    \begin{center}
    \begin{small}
    \begin{sc}
    \begin{tabular}{l @{\hskip 1.5em} cc @{\hskip 1.5em} cccc}
    \toprule
    & \multicolumn{2}{c}{3DShapes}
    & \multicolumn{4}{c}{CelebA}\\
    \midrule
    & DCI $\uparrow$ & Image Qual. & TAD$\uparrow$ & Attrs$\uparrow$ & Latent Qual. $\uparrow$ & FID $\downarrow$ \\
    \midrule
    AE & $0.219 \pm 0.001$ & \ourxmark & $0.042 \pm 0.004$ & $1.0 \pm 0.0$ & $0.759 \pm 0.003$ & $90.4 \pm 1.8$\\
    VAE & $0.276 \pm 0.001$ & \ourxmark & $0.000 \pm 0.000$ & $0.0 \pm 0.0$ & $0.770 \pm 0.002$  & $94.3 \pm 2.8$\\
    beta-VAE & \secondplace ${0.281 \pm 0.001}$ & \ourxmark & $0.088 \pm 0.051$ & $1.6 \pm 0.8$ & $0.699\pm 0.001$ & $99.8 \pm 2.4$\\
    InfoVAE & $0.134 \pm 0.001$ & \ourxmark & $ 0.000 \pm 0.000$ & $ 0.0 \pm 0.0$ & $0.757\pm 0.003$ & $77.8 \pm 1.6$\\
    DiffAE & $0.177 \pm 0.001$ & \ourcheckmark &  \secondplace$0.114 \pm 0.033$ & \secondplace$2.8 \pm 0.4$ & \secondplace$0.817 \pm 0.001$ & $8.4 \pm 0.1$\\
    \midrule
    InfoDiffusion ($\lambda=0.1,\zeta=0.9$)  & \firstplace$0.296 \pm 0.057$ & \ourxmark & \firstplace$0.439 \pm 0.134$ & \firstplace$5.7 \pm 0.6$ & \firstplace$0.844 \pm 0.005$ & $12.1 \pm 0.2$\\
    ~~ w/ guidance ($w=2$) & & & & & & $11.1 \pm 0.3$\\
    ~~ w/Learned Latent & & \ourcheckmark & & & & \secondplace$7.6 \pm 0.2$\\
    ~~ w/Learned Latent, $w=2$ & & & & & & \firstplace$6.3 \pm 0.2$\\
    \bottomrule
    \end{tabular}
    \end{sc}
    \end{small}
    \end{center}
\end{table*}

\subsection{Discrete Latent Priors}
We demonstrate the flexibility of our model by training with a discrete prior.
We train InfoDiffusion with FSQ latents (\Cref{subsec:discrete}; $d_z=32$ dimensions with $L=5$ levels each, i.e., a uniform prior over $5^{32}$ codes) on the CelebA dataset
and find that latent space quality is comparable to that of continuous latents, with average AUROC of $0.849 \pm 0.003$, a TAD score of $0.370 \pm 0.240$, and $5.6$ captured attributes over three seeds (compared to $0.844$, $0.439$, and $5.7$ for continuous InfoDiffusion latents and AUROCs of 0.70--0.82 for the continuous-latent baselines in \Cref{tab:disentangle}) (details in \Cref{app:discrete}).
\begin{table}[ht]
\setlength{\tabcolsep}{2pt}
    \caption{Representation learning comparison to contrastive methods (InfoDiffusion: $\lambda=0.1$, $\zeta=0.9$; mean $\pm$ standard deviation over three training seeds).
    `Gen.' indicates whether the model has generative capabilities.
    `Dim.' denotes the latent dimension.
    Disentanglement is measured by TAD.
    `Latent Quality' is measured as AUROC scores averaged across CelebA attributes for logistic regression classifiers trained on latent representations.
    We report means $\pm$ one standard deviation for quantitative metrics.
    Darkly shaded cells indicate the best while lightly shaded cells indicate the second best.
    $^\dagger$ denotes that the weights are taken from the PyTorch repository for the method.}
    \newcommand{\greencheck}{{\bf \color{OliveGreen}\checkmark}}

    \newcommand*\colourcheck[1]{%
      \expandafter\newcommand\csname #1check\endcsname{\textcolor{#1}{\ding{52}}}%
    }
    \definecolor{bloodred}{HTML}{B00000}
    \definecolor{cautionyellow}{HTML}{EED202}
    \newcommand*\colourxmark[1]{%
      \expandafter\newcommand\csname #1xmark\endcsname{\textcolor{#1}{\ding{54}}}%
    }
    \newcommand*\colourcheckodd[1]{%
      \expandafter\newcommand\csname #1checkodd\endcsname{\textcolor{#1}{\ding{51}}}%
    }
    \colourcheckodd{cautionyellow}
    \colourcheck{cautionyellow}
    \colourcheck{OliveGreen}
    \colourxmark{bloodred}

    \newcommand{\ourxmark}{\bloodredxmark}%
    \newcommand{\ourcheckmark}{\OliveGreencheck}

    \label{tab:contrast1}
    \centering
    \begin{small}
    \begin{sc}
    \scalebox{0.98}{
    \begin{tabular}{lcccc }
    \toprule
    CelebA & Gen. & Dim. & TAD $\uparrow$ & \makecell{Latent \\Qual. $\uparrow$} \\
    \midrule    SIMCLR$^{\dagger}$\footnotetext{The weights are taken from the respective PyTorch repositories of these models.}& \ourxmark & 2048 & $ {0.192 \pm 0.015}$ & $0.812 \pm 0.003$ \\
    MOCO-v2$^{\dagger}$ & \ourxmark & 2048 & \secondplace${0.279 \pm 0.025}$ & \firstplace${0.846 \pm 0.001}$ \\
    DINO$^{\dagger}$ & \ourxmark & 384 & $ {0.000 \pm 0.000}$ & $0.592 \pm 0.003$ \\
    InfoDiffusion & \ourcheckmark & 32 & \firstplace$0.439 \pm 0.134$ & \secondplace$0.844 \pm 0.005$ \\
    \bottomrule
    \end{tabular}}
    \end{sc}
    \end{small}
\end{table}

\begin{table}[ht!]
\setlength{\tabcolsep}{2pt}
    \caption{Representation learning comparison to SIMCLR and PDAE with 32-dimensional latents (InfoDiffusion: $\lambda=0.1$, $\zeta=0.9$; mean $\pm$ standard deviation over three training seeds).
    `Gen.' indicates whether the model has generative capabilities.
    `Attrs.' counts the number of ``captured'' attributes when calculating the TAD score.
    `Latent Quality' is measured as AUROC scores averaged across attributes for logistic regression classifiers trained on $\bz$.
    We report means $\pm$ one standard deviation for quantitative metrics.
    Darkly shaded cells indicate the best while lightly shaded cells indicate the second best.
    $^*$ denotes that the weights are taken from the PyTorch repository and fine-tuned with an added dense layer.
    $^{\ddagger}$ denotes that the model is trained by us with the codebase provided by this baseline, in earlier runs (\Cref{app:exp_details}).}
    \newcommand{\greencheck}{{\bf \color{OliveGreen}\checkmark}}

    \newcommand*\colourcheck[1]{%
      \expandafter\newcommand\csname #1check\endcsname{\textcolor{#1}{\ding{52}}}%
    }
    \definecolor{bloodred}{HTML}{B00000}
    \definecolor{cautionyellow}{HTML}{EED202}
    \newcommand*\colourxmark[1]{%
      \expandafter\newcommand\csname #1xmark\endcsname{\textcolor{#1}{\ding{54}}}%
    }
    \newcommand*\colourcheckodd[1]{%
      \expandafter\newcommand\csname #1checkodd\endcsname{\textcolor{#1}{\ding{51}}}%
    }
    \colourcheckodd{cautionyellow}
    \colourcheck{cautionyellow}
    \colourcheck{OliveGreen}
    \colourxmark{bloodred}

    \newcommand{\ourxmark}{\bloodredxmark}%
    \newcommand{\ourcheckmark}{\OliveGreencheck}

    \label{tab:contrast2}
    \centering
    \begin{small}
    \begin{sc}
    \scalebox{0.98}{
    \begin{tabular}{lcccc }
    \toprule
    CelebA & Gen. & TAD $\uparrow$ & ATTRS $\uparrow$ & \makecell{Latent \\Qual. $\uparrow$}\\
    \midrule
    SIMCLR$^*$ & \ourxmark  & $ \secondplace{0.062 \pm 0.005}$ & \secondplace$2.6 \pm 0.5$ & $0.757 \pm 0.002$ \\
    PDAE$^{\ddagger}$ & \ourcheckmark  & $ {0.009 \pm 0.001}$ & $1.0 \pm 0.0$ & \secondplace$0.767 \pm 0.003$ \\
    InfoDiff. & \ourcheckmark  & \firstplace $0.439 \pm 0.134$ & \firstplace $5.7 \pm 0.6$& \firstplace $0.844 \pm 0.005$ \\
    \bottomrule
    \end{tabular}}
    \end{sc}
    \end{small}
\end{table}

\begin{table*}[ht]
    \setlength{\tabcolsep}{5pt} %
    \caption{Analogy between progress in the space of auto-encoders and similar progress for diffusion models.}
    \label{analogy}
    \begin{small}
        \begin{tabular}{cccc}
            \toprule
            \textbf{Method} & \textbf{Non-Probabilistic} & \textbf{Probabilistic Extension} & \textbf{Regularized Extension} \\
            \midrule
            Auto-encoders & AE \cite{lecun1987phd} & VAE \cite{kingma2013auto} & InfoVAE \cite{zhao2017infovae} \\
            \midrule
            Diffusion models & DiffAE \cite{preechakul2022diffusion} &  Variational Auxiliary-Variable Diffusion \hyperref[sec:4]{Sec. 4} & InfoDiffusion \hyperref[sec:5]{Sec. 5} \\
            \bottomrule
        \end{tabular}
    \end{small}
\end{table*}

\subsection{Comparison to Contrastive Methods}
We compare the quality of our learned representations to those from established contrastive learning methods, including SimCLR \cite{chen2020simple}, MOCO-v2 \cite{chen2020improved}, and DINO \cite{caron2021emerging}.
In \Cref{tab:contrast1}, we report average AUROC for classifiers trained on $\bz$ to predict CelebA annotations and the TAD scores for disentanglement\footnote{We excluded the ``Number of attributes captured'' metric for this comparison, as the pre-trained contrastive method baselines use larger latent dimension, which artificially inflates the value for this metric.}.
Our findings indicate that our latent representations are comparable, and in some instances superior, to these robust baselines.
Our approach also has the added benefit of being a generative model.
We also note that our model uses a much smaller capacity latent variable compared to these contrastive method baselines.

When comparing to methods with similar latent dimension, InfoDiffusion is able to substantially outperform baseline models.
In \Cref{tab:contrast2}, we compare to a fine-tuned, pre-trained encoder of SIMCLR with an additional dense layer that projects to 32 dimensions.
We also introduce an another baseline, PDAE \cite{zhang2022unsupervised}, which builds an auto-encoder based on pre-trained diffusion models.
Our method outperforms these alternatives on both the disentanglement and latent quality metrics.

\subsection{Exploring InfoDiffusion Modeling Choices}\label{subsec:choices}
\paragraph{Regularization Coefficients}
An evaluation of various $\zeta$ and $\lambda$ parameters for InfoDiffusion is presented in \Cref{app:coeffs}.
Values $\zeta>1$ make the coefficient of the per-sample KL term negative, so the objective becomes unbounded (it can be increased indefinitely by shrinking the posterior variance), and \Cref{prop:divergence} requires $\zeta\le 1$; thus, we cap $\zeta = 1$.
Among the settings we evaluate, $(\lambda,\zeta)=(0.1,0.9)$, which keeps a small per-sample KL term and no MMD term, gives the most disentangled CelebA latents and the best or second-best FID on all four natural image datasets, with both samplers; its latents are less informative than those of $\zeta=1$ with small $\lambda$ on CIFAR10 and CelebA. We therefore use it in our main experiments. With $\zeta=1$, increasing $\lambda$ trades latent information for prior matching, and $\lambda=0.1$ removes the attribute information from the CelebA latents.
\vspace{-0.3cm}
\paragraph{Sampling Method}
\Cref{tab:class} and \Cref{tab:disentangle} provide results for generation using samples extracted from either the prior distribution or a learned latent distribution, as denoted in the "w/Learned Latent" rows.
Decoding $\bz\sim p(\bz)$ gives FIDs of $22.5$, $36.2$, $19.8$, and $12.1$ on FashionMNIST, CIFAR10, FFHQ, and CelebA. Sampling $\bz$ from the learned latent prior lowers them to $8.1$, $25.7$, $13.1$, and $7.6$, and guidance on $\bz$ ($w=2$) improves FID further with both samplers on all four datasets. The encodings are more concentrated than the prior (average standard deviation $0.26$--$0.56$ per dimension for $d_z=256$), so prior draws often fall in regions the decoder has rarely seen. The regularization coefficients strongly affect prior sampling: with $(\lambda,\zeta)=(0.01,1)$, the FID with $\bz\sim p(\bz)$ rises to $38$--$64$ (\Cref{app:coeffs}). Like DiffAE \citep{preechakul2022diffusion}, our model therefore obtains its best samples from a learned latent prior.

\section{Related Work}\label{sec:related}
\subsection{Representation Learning in Generative Modeling}
VAEs \cite{kingma2013auto, higgins2017beta} extend the auto-encoder framework through variational inference algorithms to produce a generative model with semantically meaningful and smooth latent spaces.
InfoVAE~\cite{zhao2017infovae} addresses a key failure mode of VAEs through mutual information regularization to improve the quality of the variational posterior, and Wasserstein auto-encoders \citep{tolstikhin2018wasserstein} replace the per-sample KL term with a divergence (e.g., MMD) between the aggregate posterior and the prior.
Another paradigm, known as Infogan
\cite{chen2016infogan}, extends generative adversarial networks (GANs; \citet{goodfellow2020generative}) by similarly using information maximization.
Our approach has the advantage of combining the stable training and generation quality of diffusion models with the representation learning capabilities of these prior works.

\subsection{Diffusion Models for Representation Learning}
Our work builds upon advances in diffusion models, which enable stable, high-resolution training on varied datasets \cite{dhariwal2021diffusion,ho2020denoising,saharia2022photorealistic,rombach2021highresolution}.
Recent work has combined auto-encoders with diffusion models---e.g., DiffAE \cite{preechakul2022diffusion}, a non-probabilistic auto-encoder model that produces semantically meaningful latents.

The relationship between our method and DiffAE is analogous to the relationship between InfoVAE \citep{zhao2017infovae} and a regular non-probabilistic auto-encoder.
Our method augments DiffAE with: (1) a principled probabilistic auxiliary-variable model family and (2) new learning objectives based on variational mutual information maximization.
This yields a number of advantages.
First, our method allows users to specify domain knowledge through a prior and supports the use of discrete variables.
Additionally, our objective removes the penalty that the ELBO places on mutual information and matches the aggregate posterior to the prior, which empirically yields more disentangled latents.

\Cref{analogy} illustrates how our approach relates to previous work on both diffusion models and mutual information regularization by showing an analogy between progress in the space of auto-encoders and similar progress for diffusion models.

\paragraph{Subsequent work} Since the original version of this paper, diffusion-based representation learning has developed further \citep{fuest2026diffusion}. SODA \citep{hudson2024soda} trains an encoder whose bottleneck latent guides a diffusion decoder that generates related novel views, and reports strong linear-probe performance on ImageNet. DisDiff \citep{yang2023disdiff} discovers factors of variation from a pre-trained diffusion model by decomposing its gradient field, and EncDiff \citep{yang2024diffusion} conditions a latent diffusion model on encoded concept tokens through cross-attention to obtain disentangled representations. RCG \citep{li2024return} shows that sampling a learned representation and then generating images conditioned on it, with guidance, substantially improves unconditional generation; this two-stage procedure is closely related to our learned-latent sampler (\Cref{app:sampling_learned}).

\section{Conclusion}\label{sec:conclusion}
In this work, we proposed InfoDiffusion, a new learning algorithm based on a diffusion model that uses an auxiliary variable to encode semantically meaningful information.
We derive InfoDiffusion from a principled probabilistic extension of diffusion models that relies on variational inference to discover low-dimensional latents.
Augmenting this variational auxiliary-variable diffusion framework with mutual information regularization enables InfoDiffusion to simultaneously achieve high-quality sample generation \emph{and} informative latent representations, which we use to control generation and improve downstream prediction.

We evaluate InfoDiffusion on several image datasets and against state-of-the-art generative and representation learning baselines and show that it produces semantically rich and more disentangled latent representations and high-quality images.
We expect InfoDiffusion will be useful in generative design and other applications that require both exploring a latent space and quality generation.

\paragraph{Limitations} Our experiments use small models at $32\times 32$ and $64\times 64$ resolution trained for 50 epochs (5 on 3DShapes), and FID is computed from 10{,}000 samples, so absolute sample quality is far from that of large-scale diffusion models. Variational lower bounds on the mutual information under the generative model, as in InfoGAN \citep{chen2016infogan}, are a natural extension. A strong diffusion decoder can still synthesize local, high-frequency attributes from $\bx_t$ alone, and the low-dimensional latents tend to omit them. Some architectural choices, such as the pooling in the encoder, the crops of the encoder input, and the distribution of training time steps, depend on the dataset (\Cref{tab:exp_details}). Finally, high-quality sampling currently relies on a learned latent prior rather than on the prior $p(\bz)$ itself.

\section*{Acknowledgements}\label{sec:ack}
This work was supported by Tata Consulting Services, the Presidential Life Science Fellowship, the Hal \& Inge Marcus PhD Fellowship, and NSF CAREER grants (\#1750326, \#2046760, and \#2145577).

\bibliography{icml}

\begin{thebibliography}{52}
\providecommand{\natexlab}[1]{#1}
\providecommand{\url}[1]{\texttt{#1}}
\expandafter\ifx\csname urlstyle\endcsname\relax
  \providecommand{\doi}[1]{doi: #1}\else
  \providecommand{\doi}{doi: \begingroup \urlstyle{rm}\Url}\fi

\bibitem[Burgess \& Kim(2018)Burgess and Kim]{3dshapes18}
Burgess, C. and Kim, H.
\newblock 3d shapes dataset.
\newblock https://github.com/deepmind/3dshapes-dataset/, 2018.

\bibitem[Caron et~al.(2021)Caron, Touvron, Misra, J{\'e}gou, Mairal,
  Bojanowski, and Joulin]{caron2021emerging}
Caron, M., Touvron, H., Misra, I., J{\'e}gou, H., Mairal, J., Bojanowski, P.,
  and Joulin, A.
\newblock Emerging properties in self-supervised vision transformers.
\newblock In \emph{Proceedings of the IEEE/CVF international conference on
  computer vision}, pp.\  9650--9660, 2021.

\bibitem[Chen et~al.(2020{\natexlab{a}})Chen, Kornblith, Norouzi, and
  Hinton]{chen2020simple}
Chen, T., Kornblith, S., Norouzi, M., and Hinton, G.
\newblock A simple framework for contrastive learning of visual
  representations.
\newblock In \emph{International conference on machine learning}, pp.\
  1597--1607. PMLR, 2020{\natexlab{a}}.

\bibitem[Chen et~al.(2016)Chen, Duan, Houthooft, Schulman, Sutskever, and
  Abbeel]{chen2016infogan}
Chen, X., Duan, Y., Houthooft, R., Schulman, J., Sutskever, I., and Abbeel, P.
\newblock Infogan: Interpretable representation learning by information
  maximizing generative adversarial nets.
\newblock \emph{Advances in neural information processing systems}, 29, 2016.

\bibitem[Chen et~al.(2020{\natexlab{b}})Chen, Fan, Girshick, and
  He]{chen2020improved}
Chen, X., Fan, H., Girshick, R., and He, K.
\newblock Improved baselines with momentum contrastive learning.
\newblock \emph{arXiv preprint arXiv:2003.04297}, 2020{\natexlab{b}}.

\bibitem[Dhariwal \& Nichol(2021)Dhariwal and Nichol]{dhariwal2021diffusion}
Dhariwal, P. and Nichol, A.
\newblock Diffusion models beat gans on image synthesis.
\newblock \emph{Advances in Neural Information Processing Systems},
  34:\penalty0 8780--8794, 2021.

\bibitem[Eastwood \& Williams(2018)Eastwood and
  Williams]{eastwood2018framework}
Eastwood, C. and Williams, C.~K.
\newblock A framework for the quantitative evaluation of disentangled
  representations.
\newblock In \emph{International Conference on Learning Representations}, 2018.

\bibitem[Esser et~al.(2024)Esser, Kulal, Blattmann, Entezari, M{\"u}ller,
  Saini, Levi, Lorenz, Sauer, Boesel, Podell, Dockhorn, English, and
  Rombach]{esser2024scaling}
Esser, P., Kulal, S., Blattmann, A., Entezari, R., M{\"u}ller, J., Saini, H.,
  Levi, Y., Lorenz, D., Sauer, A., Boesel, F., Podell, D., Dockhorn, T.,
  English, Z., and Rombach, R.
\newblock Scaling rectified flow transformers for high-resolution image
  synthesis.
\newblock In \emph{Proceedings of the 41st International Conference on Machine
  Learning}, volume 235 of \emph{Proceedings of Machine Learning Research},
  pp.\  12606--12633. PMLR, 2024.

\bibitem[Fuest et~al.(2026)Fuest, Ma, Gui, Schusterbauer, Hu, and
  Ommer]{fuest2026diffusion}
Fuest, M., Ma, P., Gui, M., Schusterbauer, J., Hu, V.~T., and Ommer, B.
\newblock Diffusion models and representation learning: A survey.
\newblock \emph{IEEE Transactions on Pattern Analysis and Machine
  Intelligence}, 48\penalty0 (7):\penalty0 7209--7228, 2026.

\bibitem[Goodfellow et~al.(2020)Goodfellow, Pouget-Abadie, Mirza, Xu,
  Warde-Farley, Ozair, Courville, and Bengio]{goodfellow2020generative}
Goodfellow, I., Pouget-Abadie, J., Mirza, M., Xu, B., Warde-Farley, D., Ozair,
  S., Courville, A., and Bengio, Y.
\newblock Generative adversarial networks.
\newblock \emph{Communications of the ACM}, 63\penalty0 (11):\penalty0
  139--144, 2020.

\bibitem[Gretton et~al.(2012)Gretton, Borgwardt, Rasch, Sch{\"o}lkopf, and
  Smola]{gretton2012kernel}
Gretton, A., Borgwardt, K.~M., Rasch, M.~J., Sch{\"o}lkopf, B., and Smola, A.
\newblock A kernel two-sample test.
\newblock \emph{The Journal of Machine Learning Research}, 13\penalty0
  (1):\penalty0 723--773, 2012.

\bibitem[Higgins et~al.(2017)Higgins, Matthey, Pal, Burgess, Glorot, Botvinick,
  Mohamed, and Lerchner]{higgins2017beta}
Higgins, I., Matthey, L., Pal, A., Burgess, C., Glorot, X., Botvinick, M.,
  Mohamed, S., and Lerchner, A.
\newblock beta-vae: Learning basic visual concepts with a constrained
  variational framework.
\newblock In \emph{International conference on learning representations}, 2017.

\bibitem[Ho \& Salimans(2022)Ho and Salimans]{ho2022classifier}
Ho, J. and Salimans, T.
\newblock Classifier-free diffusion guidance.
\newblock \emph{arXiv preprint arXiv:2207.12598}, 2022.

\bibitem[Ho et~al.(2020)Ho, Jain, and Abbeel]{ho2020denoising}
Ho, J., Jain, A., and Abbeel, P.
\newblock Denoising diffusion probabilistic models.
\newblock \emph{Advances in Neural Information Processing Systems},
  33:\penalty0 6840--6851, 2020.

\bibitem[Hudson et~al.(2024)Hudson, Zoran, Malinowski, Lampinen, Jaegle,
  McClelland, Matthey, Hill, and Lerchner]{hudson2024soda}
Hudson, D.~A., Zoran, D., Malinowski, M., Lampinen, A.~K., Jaegle, A.,
  McClelland, J.~L., Matthey, L., Hill, F., and Lerchner, A.
\newblock {SODA}: Bottleneck diffusion models for representation learning.
\newblock In \emph{Proceedings of the IEEE/CVF Conference on Computer Vision
  and Pattern Recognition (CVPR)}, pp.\  23115--23127, 2024.

\bibitem[Jang et~al.(2016)Jang, Gu, and Poole]{jang2016categorical}
Jang, E., Gu, S., and Poole, B.
\newblock Categorical reparameterization with gumbel-softmax.
\newblock \emph{arXiv preprint arXiv:1611.01144}, 2016.

\bibitem[Jing et~al.(2022)Jing, Corso, Chang, Barzilay, and
  Jaakkola]{jing2022torsional}
Jing, B., Corso, G., Chang, J., Barzilay, R., and Jaakkola, T.
\newblock Torsional diffusion for molecular conformer generation.
\newblock \emph{arXiv preprint arXiv:2206.01729}, 2022.

\bibitem[Karras et~al.(2019)Karras, Laine, and Aila]{karras2019style}
Karras, T., Laine, S., and Aila, T.
\newblock A style-based generator architecture for generative adversarial
  networks.
\newblock In \emph{Proceedings of the IEEE/CVF conference on computer vision
  and pattern recognition}, pp.\  4401--4410, 2019.

\bibitem[Karras et~al.(2022)Karras, Aittala, Aila, and
  Laine]{karras2022elucidating}
Karras, T., Aittala, M., Aila, T., and Laine, S.
\newblock Elucidating the design space of diffusion-based generative models.
\newblock In \emph{Advances in Neural Information Processing Systems},
  volume~35, pp.\  26565--26577, 2022.

\bibitem[Kingma \& Gao(2023)Kingma and Gao]{kingma2023understanding}
Kingma, D.~P. and Gao, R.
\newblock Understanding diffusion objectives as the {ELBO} with simple data
  augmentation.
\newblock In \emph{Advances in Neural Information Processing Systems},
  volume~36, pp.\  65484--65516. Curran Associates, Inc., 2023.

\bibitem[Kingma \& Welling(2013)Kingma and Welling]{kingma2013auto}
Kingma, D.~P. and Welling, M.
\newblock Auto-encoding variational bayes.
\newblock \emph{arXiv preprint arXiv:1312.6114}, 2013.

\bibitem[Kong et~al.(2020)Kong, Ping, Huang, Zhao, and
  Catanzaro]{kong2020diffwave}
Kong, Z., Ping, W., Huang, J., Zhao, K., and Catanzaro, B.
\newblock Diffwave: A versatile diffusion model for audio synthesis.
\newblock \emph{arXiv preprint arXiv:2009.09761}, 2020.

\bibitem[Krizhevsky et~al.(2009)Krizhevsky, Hinton,
  et~al.]{krizhevsky2009learning}
Krizhevsky, A., Hinton, G., et~al.
\newblock Learning multiple layers of features from tiny images.
\newblock 2009.

\bibitem[LeCun(1987)]{lecun1987phd}
LeCun, Y.
\newblock Phd thesis: Modeles connexionnistes de l'apprentissage (connectionist
  learning models).
\newblock 1987.

\bibitem[Li et~al.(2024)Li, Katabi, and He]{li2024return}
Li, T., Katabi, D., and He, K.
\newblock Return of unconditional generation: A self-supervised representation
  generation method.
\newblock In \emph{Advances in Neural Information Processing Systems},
  volume~37, pp.\  125441--125468. Curran Associates, Inc., 2024.

\bibitem[Lipman et~al.(2023)Lipman, Chen, Ben-Hamu, Nickel, and
  Le]{lipman2023flow}
Lipman, Y., Chen, R. T.~Q., Ben-Hamu, H., Nickel, M., and Le, M.
\newblock Flow matching for generative modeling.
\newblock In \emph{International Conference on Learning Representations}, 2023.

\bibitem[Liu et~al.(2023)Liu, Gong, and Liu]{liu2023flow}
Liu, X., Gong, C., and Liu, Q.
\newblock Flow straight and fast: Learning to generate and transfer data with
  rectified flow.
\newblock In \emph{International Conference on Learning Representations}, 2023.

\bibitem[Liu et~al.(2015)Liu, Luo, Wang, and Tang]{liu2015faceattributes}
Liu, Z., Luo, P., Wang, X., and Tang, X.
\newblock Deep learning face attributes in the wild.
\newblock In \emph{Proceedings of International Conference on Computer Vision
  (ICCV)}, December 2015.

\bibitem[Locatello et~al.(2019)Locatello, Bauer, Lucic, Raetsch, Gelly,
  Sch{\"o}lkopf, and Bachem]{locatello2019challenging}
Locatello, F., Bauer, S., Lucic, M., Raetsch, G., Gelly, S., Sch{\"o}lkopf, B.,
  and Bachem, O.
\newblock Challenging common assumptions in the unsupervised learning of
  disentangled representations.
\newblock In \emph{international conference on machine learning}, pp.\
  4114--4124. PMLR, 2019.

\bibitem[Maddison et~al.(2017)Maddison, Mnih, and Teh]{maddison2017concrete}
Maddison, C.~J., Mnih, A., and Teh, Y.~W.
\newblock The concrete distribution: A continuous relaxation of discrete random
  variables.
\newblock In \emph{International Conference on Learning Representations}, 2017.

\bibitem[Makhzani et~al.(2015)Makhzani, Shlens, Jaitly, Goodfellow, and
  Frey]{makhzani2015adversarial}
Makhzani, A., Shlens, J., Jaitly, N., Goodfellow, I., and Frey, B.
\newblock Adversarial autoencoders.
\newblock \emph{arXiv preprint arXiv:1511.05644}, 2015.

\bibitem[Mentzer et~al.(2024)Mentzer, Minnen, Agustsson, and
  Tschannen]{mentzer2024finite}
Mentzer, F., Minnen, D., Agustsson, E., and Tschannen, M.
\newblock Finite scalar quantization: {VQ-VAE} made simple.
\newblock In \emph{International Conference on Learning Representations}, 2024.

\bibitem[Parmar et~al.(2022)Parmar, Zhang, and Zhu]{parmar2022aliased}
Parmar, G., Zhang, R., and Zhu, J.-Y.
\newblock On aliased resizing and surprising subtleties in {GAN} evaluation.
\newblock In \emph{Proceedings of the IEEE/CVF Conference on Computer Vision
  and Pattern Recognition (CVPR)}, pp.\  11410--11420, 2022.

\bibitem[Paszke et~al.(2019)Paszke, Gross, Massa, Lerer, Bradbury, Chanan,
  Killeen, Lin, Gimelshein, Antiga, Desmaison, Kopf, Yang, DeVito, Raison,
  Tejani, Chilamkurthy, Steiner, Fang, Bai, and Chintala]{paszke2019pytorch}
Paszke, A., Gross, S., Massa, F., Lerer, A., Bradbury, J., Chanan, G., Killeen,
  T., Lin, Z., Gimelshein, N., Antiga, L., Desmaison, A., Kopf, A., Yang, E.,
  DeVito, Z., Raison, M., Tejani, A., Chilamkurthy, S., Steiner, B., Fang, L.,
  Bai, J., and Chintala, S.
\newblock Pytorch: An imperative style, high-performance deep learning library.
\newblock In Wallach, H., Larochelle, H., Beygelzimer, A., d'Alch\'{e} Buc, F.,
  Fox, E., and Garnett, R. (eds.), \emph{Advances in Neural Information
  Processing Systems 32}, pp.\  8024--8035. Curran Associates, Inc., 2019.

\bibitem[Pedregosa et~al.(2011)Pedregosa, Varoquaux, Gramfort, Michel, Thirion,
  Grisel, Blondel, Prettenhofer, Weiss, Dubourg, Vanderplas, Passos,
  Cournapeau, Brucher, Perrot, and Duchesnay]{scikit-learn}
Pedregosa, F., Varoquaux, G., Gramfort, A., Michel, V., Thirion, B., Grisel,
  O., Blondel, M., Prettenhofer, P., Weiss, R., Dubourg, V., Vanderplas, J.,
  Passos, A., Cournapeau, D., Brucher, M., Perrot, M., and Duchesnay, E.
\newblock Scikit-learn: Machine learning in {P}ython.
\newblock \emph{Journal of Machine Learning Research}, 12:\penalty0 2825--2830,
  2011.

\bibitem[Preechakul et~al.(2022)Preechakul, Chatthee, Wizadwongsa, and
  Suwajanakorn]{preechakul2022diffusion}
Preechakul, K., Chatthee, N., Wizadwongsa, S., and Suwajanakorn, S.
\newblock Diffusion autoencoders: Toward a meaningful and decodable
  representation.
\newblock In \emph{Proceedings of the IEEE/CVF Conference on Computer Vision
  and Pattern Recognition}, pp.\  10619--10629, 2022.

\bibitem[Ramesh et~al.(2022)Ramesh, Dhariwal, Nichol, Chu, and
  Chen]{ramesh2022hierarchical}
Ramesh, A., Dhariwal, P., Nichol, A., Chu, C., and Chen, M.
\newblock Hierarchical text-conditional image generation with clip latents.
\newblock \emph{arXiv preprint arXiv:2204.06125}, 2022.

\bibitem[Rombach et~al.(2021)Rombach, Blattmann, Lorenz, Esser, and
  Ommer]{rombach2021highresolution}
Rombach, R., Blattmann, A., Lorenz, D., Esser, P., and Ommer, B.
\newblock High-resolution image synthesis with latent diffusion models, 2021.

\bibitem[Ronneberger et~al.(2015)Ronneberger, Fischer, and
  Brox]{ronneberger2015u}
Ronneberger, O., Fischer, P., and Brox, T.
\newblock U-net: Convolutional networks for biomedical image segmentation.
\newblock In \emph{Medical Image Computing and Computer-Assisted
  Intervention--MICCAI 2015: 18th International Conference, Munich, Germany,
  October 5-9, 2015, Proceedings, Part III 18}, pp.\  234--241. Springer, 2015.

\bibitem[Saharia et~al.(2022)Saharia, Chan, Saxena, Li, Whang, Denton,
  Ghasemipour, Ayan, Mahdavi, Lopes, et~al.]{saharia2022photorealistic}
Saharia, C., Chan, W., Saxena, S., Li, L., Whang, J., Denton, E., Ghasemipour,
  S. K.~S., Ayan, B.~K., Mahdavi, S.~S., Lopes, R.~G., et~al.
\newblock Photorealistic text-to-image diffusion models with deep language
  understanding.
\newblock \emph{arXiv preprint arXiv:2205.11487}, 2022.

\bibitem[Song et~al.(2021)Song, Meng, and Ermon]{song2021denoising}
Song, J., Meng, C., and Ermon, S.
\newblock Denoising diffusion implicit models.
\newblock In \emph{International Conference on Learning Representations}, 2021.

\bibitem[Tolstikhin et~al.(2018)Tolstikhin, Bousquet, Gelly, and
  Sch{\"o}lkopf]{tolstikhin2018wasserstein}
Tolstikhin, I., Bousquet, O., Gelly, S., and Sch{\"o}lkopf, B.
\newblock {Wasserstein} auto-encoders.
\newblock In \emph{International Conference on Learning Representations}, 2018.

\bibitem[Van Den~Oord et~al.(2017)Van Den~Oord, Vinyals, et~al.]{van2017neural}
Van Den~Oord, A., Vinyals, O., et~al.
\newblock Neural discrete representation learning.
\newblock \emph{Advances in neural information processing systems}, 30, 2017.

\bibitem[Xiao et~al.(2017)Xiao, Rasul, and Vollgraf]{xiao2017fashion}
Xiao, H., Rasul, K., and Vollgraf, R.
\newblock Fashion-mnist: a novel image dataset for benchmarking machine
  learning algorithms.
\newblock \emph{arXiv preprint arXiv:1708.07747}, 2017.

\bibitem[Xu et~al.(2022)Xu, Yu, Song, Shi, Ermon, and Tang]{xu2022geodiff}
Xu, M., Yu, L., Song, Y., Shi, C., Ermon, S., and Tang, J.
\newblock Geodiff: A geometric diffusion model for molecular conformation
  generation.
\newblock \emph{arXiv preprint arXiv:2203.02923}, 2022.

\bibitem[Yang et~al.(2022)Yang, Zhang, Song, Hong, Xu, Zhao, Shao, Zhang, Cui,
  and Yang]{yang2022diffusion}
Yang, L., Zhang, Z., Song, Y., Hong, S., Xu, R., Zhao, Y., Shao, Y., Zhang, W.,
  Cui, B., and Yang, M.-H.
\newblock Diffusion models: A comprehensive survey of methods and applications.
\newblock \emph{arXiv preprint arXiv:2209.00796}, 2022.

\bibitem[Yang et~al.(2023)Yang, Wang, Lu, and Zheng]{yang2023disdiff}
Yang, T., Wang, Y., Lu, Y., and Zheng, N.
\newblock {DisDiff}: Unsupervised disentanglement of diffusion probabilistic
  models.
\newblock In \emph{Advances in Neural Information Processing Systems},
  volume~36, pp.\  69130--69156. Curran Associates, Inc., 2023.

\bibitem[Yang et~al.(2024)Yang, Lan, Lu, and Zheng]{yang2024diffusion}
Yang, T., Lan, C., Lu, Y., and Zheng, N.
\newblock Diffusion model with cross attention as an inductive bias for
  disentanglement.
\newblock In \emph{Advances in Neural Information Processing Systems},
  volume~37, pp.\  82465--82492. Curran Associates, Inc., 2024.

\bibitem[Yeats et~al.(2022)Yeats, Liu, Womble, and Li]{yeats2022nashae}
Yeats, E., Liu, F., Womble, D., and Li, H.
\newblock Nashae: Disentangling representations through adversarial covariance
  minimization.
\newblock In \emph{Computer Vision--ECCV 2022: 17th European Conference, Tel
  Aviv, Israel, October 23--27, 2022, Proceedings, Part XXVII}, pp.\  36--51.
  Springer, 2022.

\bibitem[Yu et~al.(2024)Yu, Lezama, Gundavarapu, Versari, Sohn, Minnen, Cheng,
  Gupta, Gu, Hauptmann, Gong, Yang, Essa, Ross, and Jiang]{yu2024language}
Yu, L., Lezama, J., Gundavarapu, N.~B., Versari, L., Sohn, K., Minnen, D.,
  Cheng, Y., Gupta, A., Gu, X., Hauptmann, A.~G., Gong, B., Yang, M.-H., Essa,
  I., Ross, D.~A., and Jiang, L.
\newblock Language model beats diffusion -- tokenizer is key to visual
  generation.
\newblock In \emph{International Conference on Learning Representations}, 2024.

\bibitem[Zhang et~al.(2022)Zhang, Zhao, and Lin]{zhang2022unsupervised}
Zhang, Z., Zhao, Z., and Lin, Z.
\newblock Unsupervised representation learning from pre-trained diffusion
  probabilistic models.
\newblock \emph{Advances in Neural Information Processing Systems},
  35:\penalty0 22117--22130, 2022.

\bibitem[Zhao et~al.(2017)Zhao, Song, and Ermon]{zhao2017infovae}
Zhao, S., Song, J., and Ermon, S.
\newblock Infovae: Information maximizing variational autoencoders.
\newblock \emph{arXiv preprint arXiv:1706.02262}, 2017.

\end{thebibliography}
\bibliographystyle{icml2023}

\newpage
\appendix
\onecolumn
\section{Proof of Proposition 5.1}\label{app:prop1}
We start with the derivation for the ELBO of a Variational Auxiliary-Variable Diffusion Model defined in \Cref{eq:denoise}:
\begingroup
\allowdisplaybreaks
\begin{align}
\log p(\bx_{0})
&= \log \int p(\bx_{0:T}, \bz) d\bx_{1:T} d\bz \nonumber\\
&= \log \int \frac{p(\bx_{0:T}, \bz)q(\bx_{1:T}\mid\bx_0)q_\phi(\bz \mid \bx_0)}{q(\bx_{1:T}\mid\bx_0)q_\phi(\bz\mid\bx_0)} d\bx_{1:T} d\bz \nonumber\\
&= \log\Exp_{q(\bx_{1:T}\mid\bx_0)}\left[\Exp_{q_\phi(\bz\mid\bx_0)}\left[\frac{p(\bx_{0:T}, \bz)}{q(\bx_{1:T}\mid\bx_0)q_\phi(\bz\mid\bx_0)}\right]\right]\nonumber\\
&\geq \Exp_{q(\bx_{1:T}\mid\bx_0)}\left[\Exp_{q_\phi(\bz\mid\bx_0)}\left[\log \frac{p(\bx_{0:T}, \bz)}{q(\bx_{1:T}\mid\bx_0)q_\phi(\bz\mid\bx_0)}\right]\right] \nonumber\\
&= \Exp_{q(\bx_{1:T}\mid\bx_0)}\left[\Exp_{q_\phi(\bz\mid\bx_0)}\left[\log \frac{p(\bz)p(\bx_T)\prod_{t=1}^{T}p(\bx_{t-1}|\bx_t, \bz)}{q_\phi(\bz\mid\bx_0)\prod_{t = 1}^{T}q(\bx_{t}|\bx_{t-1})}\right]\right] \nonumber\\
&= \Exp_{q(\bx_{1:T}\mid\bx_0)}\left[\Exp_{q_\phi(\bz\mid\bx_0)}\left[\log \frac{p(\bz)p(\bx_T)p(\bx_0|\bx_1, \bz)\prod_{t=2}^{T}p(\bx_{t-1}|\bx_t, \bz)}{q_\phi(\bz\mid\bx_0)q(\bx_1|\bx_{0})\prod_{t = 2}^{T}q(\bx_{t}|\bx_{t-1}, \bx_0)}\right]\right] \nonumber\\
&= \Exp_{q(\bx_{1:T}\mid\bx_0)}\left[\Exp_{q_\phi(\bz\mid\bx_0)}\left[\log \frac{p(\bz)}{q_\phi(\bz\mid\bx_0)} + \log \frac{p(\bx_T)p(\bx_0|\bx_1, \bz)}{q(\bx_1|\bx_0)} + \sum_{t=2}^{T}\log\frac{p(\bx_{t-1}|\bx_t, \bz)}{q(\bx_{t}|\bx_{t-1}, \bx_0)}\right]\right] \nonumber\\
&= \Exp_{q(\bx_{1:T}\mid\bx_0)}\left[\Exp_{q_\phi(\bz\mid\bx_0)}\left[\log \frac{p(\bz)}{q_\phi(\bz\mid\bx_0)} + \log \frac{p(\bx_T)p(\bx_0|\bx_1, \bz)}{q(\bx_1|\bx_0)} + \sum_{t=2}^{T}\log\frac{p(\bx_{t-1}|\bx_t, \bz)}{\frac{q(\bx_{t-1}|\bx_{t}, \bx_0)q(\bx_t|\bx_0)}{q(\bx_{t-1}|\bx_0)}}\right]\right]\nonumber\\
&= \Exp_{q(\bx_{1:T}\mid\bx_0)}\left[\Exp_{q_\phi(\bz\mid\bx_0)}\left[\log \frac{p(\bz)}{q_\phi(\bz\mid\bx_0)} + \log \frac{p(\bx_T)p(\bx_0|\bx_1, \bz)}{q(\bx_1|\bx_0)} + \log \frac{q(\bx_1|\bx_0)}{q(\bx_T|\bx_0)} +  \sum_{t=2}^{T}\log\frac{p(\bx_{t-1}|\bx_t, \bz)}{q(\bx_{t-1}|\bx_{t}, \bx_0)}\right]\right] \nonumber\\
&= \Exp_{q(\bx_{1:T}\mid\bx_0)}\left[\Exp_{q_\phi(\bz\mid\bx_0)}\left[\log \frac{p(\bz)}{q_\phi(\bz\mid\bx_0)} + \log \frac{p(\bx_T)p(\bx_0|\bx_1, \bz)}{q(\bx_T|\bx_0)} +  \sum_{t=2}^{T}\log\frac{p(\bx_{t-1}|\bx_t, \bz)}{q(\bx_{t-1}|\bx_{t}, \bx_0)}\right]\right] \nonumber\\
\begin{split}
&= \Exp_{q_\phi(\bz\mid\bx_0)}\left[\log \frac{p(\bz)}{q_\phi(\bz\mid\bx_0)}\right] + \Exp_{q(\bx_{1} \mid\bx_0)}\left[\Exp_{q_\phi(\bz\mid\bx_0)}\left[\log p_{\theta}(\bx_0|\bx_1, \bz)\right]\right] \\
&~~~~~+ \Exp_{q(\bx_{T}|\bx_0)}\left[\log \frac{p(\bx_T)}{q(\bx_T|\bx_0)}\right] + \sum_{t=2}^{T}\Exp_{q(\bx_{t-1},\bx_t \mid\bx_0)}\left[\Exp_{q_\phi(\bz\mid\bx_0)}\left[\log\frac{p(\bx_{t-1}|\bx_t, \bz)}{q(\bx_{t-1}|\bx_{t}, \bx_0)}\right]\right]
\end{split}\nonumber \\
\begin{split}\label{eq:20}
&=\Exp_{q(\bx_{1}\mid\bx_0)}\left[\Exp_{q_\phi(\bz\mid\bx_0)}\left[\log p(\bx_0\mid\bx_1, \bz)\right]\right] - \KL({q(\bx_T \mid \bx_0)} || {p(\bx_T)}) - \KL({q_\phi(\bz \mid \bx_0)} || {p(\bz)}) \\
&~~~~~- \sum_{t=2}^{T} \Exp_{q(\bx_t \mid\bx_0)}\left[\Exp_{q_\phi(\bz\mid\bx_0)}\left[\KL(q(\bx_{t-1}|\bx_t, \bx_0) || p(\bx_{t-1}\mid \bx_t, \bz))\right]\right].
\end{split}
\end{align}
\endgroup

Averaging \Cref{eq:20} over the data distribution $q(\bx_0)$, the prior matching term (the third term in \Cref{eq:20}) can be rewritten as:
\begingroup
\allowdisplaybreaks
\begin{align}
- \Exp_{q{(\bx_0)}}\KL({q_\phi(\bz | \bx_0)} || {p(\bz)})
&= \Exp_{q(\bx_0)}\left[\Exp_{q_\phi(\bz\mid\bx_0)} \left[\log p(\bz) - \log q_\phi(\bz\mid\bx_0)\right]\right] \nonumber\\
&= \Exp_{q_\phi(\bx_0, \bz)} \left[\log \frac{p(\bz)}{q_\phi(\bx_0, \bz)} + \log q(\bx_0)\right] \nonumber\\
&= \Exp_{q_\phi(\bx_0, \bz)} \left[\log \frac{p(\bz)}{q_\phi(\bx_0|\bz)q_\phi(\bz)} + \log q(\bx_0)\right] \nonumber \\
&= \Exp_{q_\phi(\bx_0, \bz)} \left[\log \frac{p(\bz)}{q_\phi(\bz)} + \log \frac{q(\bx_0)}{q_\phi(\bx_0|\bz)}\right] \nonumber \\
&= \Exp_{q_\phi(\bx_0, \bz)} \left[\log \frac{p(\bz)}{q_\phi(\bz)} + \log \frac{q_\phi(\bz)}{q_\phi(\bz|\bx_0)}\right] \nonumber \\
&= - \KL(q_\phi(\bz) || p(\bz)) - \MI_{\bx_0, \bz}.
\label{eq:25}
\end{align}
\endgroup

If we scale $\KL(q_\phi(\bz) || p(\bz))$ by $\lambda$ (i.e., we add $-\beta\,\KL(q_\phi(\bz)\|p(\bz))$ with $\beta = \lambda - 1$, as in \Cref{eq:infodiff}) and add a scaled mutual information term between $\bx_0$ and $\bz$, $\zeta \MI_{\bx_0, \bz}$, \Cref{eq:25} becomes:
\begingroup
\allowdisplaybreaks
\begin{align}
& - \lambda \KL(q_\phi(\bz) || p(\bz)) - \MI_{\bx_0, \bz} + \zeta \MI_{\bx_0, \bz} \label{eq:26} \\
=&~\Exp_{q_\phi(\bx_0, \bz)} \left[- \lambda \log \frac{q_\phi(\bz)}{p(\bz)} - (\zeta-1) \log \frac{q_\phi(\bz)}{q_\phi(\bz\mid\bx_0)} \right] \nonumber \\
=&~\Exp_{q_\phi(\bx_0, \bz)} \left[ - \log \frac{q_\phi(\bz)^{\lambda+ \zeta-1}q_\phi(\bz\mid\bx_0)^{1-\zeta}}{p(\bz)^{\lambda + \zeta-1}p(\bz)^{1-\zeta}} \right] \nonumber \\
=&~- (\lambda + \zeta-1) \KL(q_\phi(\bz) || p(\bz)) - (1-\zeta) \Exp_{q(\bx_0)} \KL(q_\phi(\bz\mid\bx_0)||p(\bz)) \label{eq:30}
\end{align}

Replacing the prior regularization term in \Cref{eq:20} with \Cref{eq:30} and averaging the remaining terms in \Cref{eq:20} over the data distribution $q(\bx_0)$, we have our InfoDiffusion ELBO objective as follows:
\begingroup
\allowdisplaybreaks
\begin{align}\label{eq:31}
\Lc_I &= \Exp_{q(\bx_0, \bx_1)}\left[\Exp_{q_\phi(\bz\mid\bx_0)}\left[\log p_{\theta}(\bx_0|\bx_1, \bz)\right]\right] - \Exp_{q(\bx_0)}[\KL({q(\bx_T | \bx_0)} || {p(\bx_T)})] \nonumber \\
&~~- \sum_{t=2}^{T} \Exp_{q(\bx_0, \bx_t)}\left[\Exp_{q_\phi(\bz\mid\bx_0)}\left[\KL({q(\bx_{t-1}|\bx_t, \bx_0)} ||{p_{\theta}(\bx_{t-1}|\bx_t, \bz)})\right]\right] \nonumber \\
&~~- (\lambda + \zeta-1) \KL(q_\phi(\bz) || p(\bz)) - (1-\zeta) \mathbb{E}_{q(\bx_0)} [\KL(q_\phi(\bz\mid\bx_0)||p(\bz))] && \square
\end{align}
We parameterize $p_\theta$ and $q_\phi$ with neural networks.

\section{Proof of Proposition 5.2}\label{app:prop2}
Following \citet{zhao2017infovae}, we first rewrite the $\Lc_I$ objective from \Cref{eq:31} with the following changes: (1) we replace the KL divergence between $q_\phi(\bz)$ and $p(\bz)$ with any strict divergence $\D$, and (2) we expand the last term of \Cref{eq:31} into a KL divergence term and a mutual information term (as in \Cref{eq:25}):
\begingroup
\allowdisplaybreaks
\begin{align}\label{eq:32}
\Lc_I &= \Exp_{q(\bx_0, \bx_1)}\left[\Exp_{q_\phi(\bz\mid\bx_0)}\left[\log p_{\theta}(\bx_0|\bx_1, \bz)\right]\right] - \Exp_{q(\bx_0)} \KL({q(\bx_T | \bx_0)} || {p(\bx_T)}) \nonumber \\
&~~- \sum_{t=2}^{T} \Exp_{q(\bx_0, \bx_t)}\left[\Exp_{q_\phi(\bz\mid\bx_0)}\left[\KL({q(\bx_{t-1}|\bx_t, \bx_0)} ||{p_{\theta}(\bx_{t-1}|\bx_t, \bz)})\right]\right] \nonumber \\
&~~- (\lambda + \zeta - 1) \D(q_\phi(\bz) || p(\bz)) - (1 - \zeta) \KL (q_\phi(\bz)||p(\bz)) - (1-\zeta)\MI_{\bx_0, \bz}.
\end{align}
Note that restricting $\zeta \leq 1$ and $\lambda \geq 1-\zeta$ (i.e., $\beta \geq -\zeta$; note that $\lambda\ge 0$ alone does not suffice), we have $1 - \zeta \geq 0$ and $\zeta + \lambda - 1 \geq 0$.
For convenience, we define
\begin{align*}
\eta &~:= ~1 - \zeta ~~\geq ~0 \\
\gamma &~:= ~\zeta + \lambda - 1 ~~\geq ~0
\end{align*}
Then, we consider the rewritten objective \Cref{eq:32} in two separate terms:
\begin{align*}
\mathcal{L}_1 &= \Exp_{q(\bx_0, \bx_1)}\left[\Exp_{q_\phi(\bz | \bx_{0})}\left[\log p_{\theta}(\bx_0|\bx_1, \bz)\right]\right] - \sum_{t=2}^{T} \Exp_{q(\bx_0, \bx_t)}\left[\Exp_{q_\phi(\bz | \bx_{0})}\left[\KL({q(\bx_{t-1}|\bx_t, \bx_0)} ||{p_{\theta}(\bx_{t-1}|\bx_t, \bz)})\right]\right] - \eta\MI_{\bx_0, \bz} \\
\mathcal{L}_2 &= - \gamma \D(q_\phi(\bz) || p(\bz)) - \eta \KL(q_\phi(\bz)||p(\bz))
\end{align*}

We will demonstrate that the two terms are maximized according to the condition in the proposition, for any values of $\eta \geq 0$ and $\gamma \geq 0$. To begin, we examine $\mathcal{L}_1$, for some fixed value of $\MI_{\bx_0, \bz} = I_0.$
Let $q_\phi(\bx_{0:T}, \bz) := q(\bx_0)\,q_\phi(\bz\mid\bx_0)\,q(\bx_{1:T}\mid\bx_0)$ denote the joint distribution induced by the data, the encoder, and the forward process, so that all expectations in $\mathcal{L}_1$ are taken under $q_\phi$. Under $q_\phi$, $\bz$ is independent of $\bx_{1:T}$ given $\bx_0$, and, because the forward process is Markov, $\bx_{t+1:T}$ is independent of $(\bx_{0:t-1}, \bz)$ given $\bx_t$; hence the reverse process is Markov given $\bz$: $q_\phi(\bx_{t-1}\mid\bx_{t:T},\bz) = q_\phi(\bx_{t-1}\mid\bx_t,\bz)$.
Writing $\KL(q(\bx_{t-1}\mid\bx_t,\bx_0)\,\|\,p_\theta(\bx_{t-1}\mid\bx_t,\bz)) = -\Hc[q(\bx_{t-1}\mid\bx_t,\bx_0)] - \Exp_{q(\bx_{t-1}\mid\bx_t,\bx_0)}[\log p_\theta(\bx_{t-1}\mid\bx_t,\bz)]$, we obtain
\begin{align*}
\mathcal{L}_1 = \sum_{t=1}^{T} \Exp_{q_\phi(\bx_{t-1},\bx_t,\bz)}\left[\log p_\theta(\bx_{t-1}\mid\bx_t,\bz)\right] + C_q - \eta I_0,
\end{align*}
where $C_q = \sum_{t=2}^T \Exp_{q(\bx_0,\bx_t)}\Hc[q(\bx_{t-1}\mid\bx_t,\bx_0)]$ depends only on $q(\bx_0)$ and the forward process. Each summand equals $-\Hc_{q_\phi}(\bx_{t-1}\mid\bx_t,\bz) - \Exp_{q_\phi(\bx_t,\bz)}\KL(q_\phi(\bx_{t-1}\mid\bx_t,\bz)\,\|\,p_\theta(\bx_{t-1}\mid\bx_t,\bz))$. Thus, for a sufficiently flexible decoder, $\mathcal{L}_1$ is maximized if and only if $p_\theta(\bx_{t-1}\mid\bx_t,\bz) = q_\phi(\bx_{t-1}\mid\bx_t,\bz)$ for all $t\ge 1$. Note that the decoder cannot match $q(\bx_{t-1}\mid\bx_t,\bx_0)$, since it does not observe $\bx_0$. At the optimum, using the reverse Markov property, the chain rule of entropy, and the conditional independence of $\bz$ and $\bx_{1:T}$ given $\bx_0$,
\begin{align*}
\sum_{t=1}^{T}\Hc_{q_\phi}(\bx_{t-1}\mid\bx_t,\bz) = \Hc_{q_\phi}(\bx_{0:T}\mid\bz) - \Hc_{q_\phi}(\bx_T\mid\bz) = \Hc_{q_\phi}(\bx_0\mid\bz) + \Hc_q(\bx_{1:T}\mid\bx_0) - \Hc_{q_\phi}(\bx_T\mid\bz).
\end{align*}
Substituting $\Hc_{q_\phi}(\bx_0\mid\bz) = \Hc_q(\bx_0) - I_0$ and $\Hc_{q_\phi}(\bx_T\mid\bz) = \Hc_q(\bx_T) - \MI(\bx_T;\bz)$ gives
\begin{align*}
\mathcal{L}^*_1 = (1-\eta)\, I_0 - \MI(\bx_T;\bz) + C'_q = \zeta I_0 - \MI(\bx_T;\bz) + C'_q,
\end{align*}
where $C'_q = C_q - \Hc_q(\bx_0) - \Hc_q(\bx_{1:T}\mid\bx_0) + \Hc_q(\bx_T)$ does not depend on $\theta$ or $\phi$. By the data processing inequality, $0 \le \MI(\bx_T;\bz) \le \MI(\bx_T;\bx_0)$, which vanishes as $\bar\alpha_T\to 0$ and is negligible for standard noise schedules. Hence, for a fixed value $I_0$ of the mutual information, $\mathcal{L}^*_1$ depends on $q_\phi$ only through $I_0$ (exactly in the limit $\bar\alpha_T \to 0$).
So we only have to independently maximize $\mathcal{L}_2,$ subject to a fixed $\MI_{\bx_0, \bz} = I_0$.

Notice that $\mathcal{L}_2$ is maximized when $q_\phi(\bz) = p(\bz)$, and thus any strict divergence $\D$ can be substituted for the $\KL$ divergence between $q_\phi(\bz)$ and $p(\bz)$, as stated in the proposition.
We thus need to show that $q_\phi(\bz) = p(\bz)$ is possible.
When $q_\phi$ is sufficiently flexible we simply have to partition the support set $\mathcal{A}$ of $p(\bz)$ into $N=e^{I_0}$ subsets (for simplicity we assume that $e^{I_0}$ is an integer, as in \citet{zhao2017infovae}) $\lbrace A_1, \cdots, A_N \rbrace$, so that each subset satisfies $\int_{A_i} p(\bz) d\bz = 1/N$.
Similarly we partition the support set $\mathcal{B}$ of $q(\bx_{0})$ into $N$ subsets $\lbrace B_1, \cdots, B_N \rbrace$, so that each subset satisfies $\int_{B_i} q(\bx_0) d\bx_0 = 1/N$.
Then we construct $q_\phi(\bz\mid\bx_0)$ mapping each $B_i$ to $A_i$ as follows
\begin{align*}
q_\phi(\bz\mid\bx_0) = \left\lbrace \begin{array}{ll} N p(\bz) & \bz \in A_i \\ 0 & \mathrm{otherwise} \end{array}
\right.
\end{align*}
for any $\bx_0 \in B_i$. It is easy to see that this distribution is normalized because
\begin{align*}
    \int_{\bz} q_\phi(\bz\mid\bx_0) d\bz = \int_{A_i} Np(\bz) d\bz = 1
\end{align*}

Then, the equality $p(\bz) = q_\phi(\bz)$ can be established through the construction of the conditional distribution $q_\phi(\bz\mid\bx_0)$. This construction is carried out in a way that, when summed or integrated over all $\bx_0$, gives us the unconditional distribution $q_\phi(\bz)$ that matches the target distribution $p(\bz)$.

Specifically, to obtain the unconditional distribution $q_\phi(\bz)$, we need to sum up over all $\bx_0$, mathematically:
\begin{align*}
q_\phi(\bz) = \int_{\bx_0} q_\phi(\bz\mid\bx_0) q(\bx_0) d\bx_0
\end{align*}

Given the way $q_\phi(\bz\mid\bx_0)$ is defined, for a particular $\bz \in A_i$, this would mean summing up $Np(\bz)$ exactly $N$ times (as we have partitioned $\mathcal{B}$ into $N$ subsets and each $\bx_0$ in a particular $B_i$ gives the same $Np(\bz)$). This will result in the equality $q_\phi(\bz) = p(\bz)$, hence demonstrating that such a match between $q_\phi(\bz)$ and $p(\bz)$ is indeed feasible. In addition,
\begin{align*}
&\MI_{\bx_0, \bz} = H_q(\bz) - H_q(\bz\mid\bx_0) \\
&= H_q(\bz) + \int_\mathcal{B} q(\bx_0) \int_{\mathcal{A}} q_\phi(\bz\mid\bx_0) \log q_\phi(\bz\mid\bx_0) d\bz d\bx_0 \\
&= H_q(\bz) + \frac{1}{N} \sum_i \int_{B_i} \int_{\mathcal{A}} q_\phi(\bz\mid\bx_0) \log q_\phi(\bz\mid\bx_0) d\bz d\bx_0 \\
&= H_q(\bz) + \frac{1}{N} \sum_i \int_{A_i} N q_\phi(\bz) \log (N q_\phi(\bz)) d\bz \\
&= H_q(\bz) + \sum_i \int_{A_i} q_\phi(\bz) \left(I_0 + \log q_\phi(\bz)\right) d\bz \\
&= H_q(\bz) + \int_{\mathcal{A}} q_\phi(\bz) \left(I_0 + \log q_\phi(\bz)\right) d\bz \\
&= H_q(\bz) + I_0 - H_q(\bz) = I_0
\end{align*}
Then we reached the maximum for both objectives
\begin{align*}
&\mathcal{L}_1^* = \zeta I_0 - \MI(\bx_T;\bz) + C'_q \\
&\mathcal{L}_2^* = 0
\end{align*}
so their sum must also be maximized. Under this optimal solution we have that $p_{\theta}(\bx_{t-1}|\bx_t, \bz) = q_\phi(\bx_{t-1}|\bx_t, \bz)$ for all $t \geq 1$ and $q_\phi(\bz) = p(\bz)$. If, in addition, $p(\bx_T) = q_\phi(\bx_T\mid\bz)$, which holds in the limit $\bar\alpha_T\to 0$, the reverse Markov factorization gives $p_\theta(\bx_{0}, \bx_{1:T}, \bz) = p(\bz)p(\bx_T)\prod_{t=1}^T p_\theta(\bx_{t-1}\mid\bx_t,\bz) = q_\phi(\bz)q_\phi(\bx_T\mid\bz)\prod_{t=1}^T q_\phi(\bx_{t-1}\mid\bx_t,\bz) = q_\phi(\bx_{0}, \bx_{1:T}, \bz)$, which implies $p_\theta(\bx_0) = q(\bx_0).$ \hfill $\square$

\section{Additional Experimental Details}\label{app:exp_details}
\paragraph{InfoDiffusion training} \Cref{tab:exp_details} lists the dataset-specific settings. All InfoDiffusion models use the continuous-time parameterization of \Cref{sec:4} and are trained with AdamW ($\beta_1=0.9$, $\beta_2=0.99$, no weight decay), batch size 128, and a learning rate of $2\times10^{-4}$ that is warmed up linearly over the first 1{,}000 steps and then cosine-annealed to $2\times10^{-5}$. Gradients are clipped to norm 1, we keep an exponential moving average (EMA) of the weights with decay $0.9995$ and use it for all evaluations, and we train in bfloat16 mixed precision. The decoder is an ADM-style U-Net \citep{dhariwal2021diffusion} with channel multipliers $(1,2,2,2)$, two residual blocks per resolution, self-attention at resolutions $16\times16$ and $8\times8$, and dropout $0.1$; the encoder uses the same multipliers (\Cref{app:arch}). We train for 50 epochs, as for the baselines, except on 3DShapes, where we train for 5 epochs. Training images are randomly flipped horizontally, except on 3DShapes. The models used to evaluate latents ($d_z=32$; $d_z=10$ on 3DShapes) are trained without latent dropout, because they are not used with guidance; the generative models ($d_z=256$) use latent dropout with probability $0.1$ (\Cref{subsec:sampling}). Unless stated otherwise, InfoDiffusion uses $\lambda=0.1$ and $\zeta=0.9$, i.e., a per-sample KL weight of $0.1$, no MMD term, and a diagonal Gaussian encoder. All models are trained on single NVIDIA H100 GPUs; we train 3 seeds per configuration for the main results and 1 seed for ablations.

\paragraph{Baselines} DiffAE is trained on every dataset with its official implementation \citep{preechakul2022diffusion}, using the official 64$\times$64 autoencoder template (\texttt{ffhq64\_autoenc}, and \texttt{celeba64d2c\_autoenc} for CelebA, which uses the same network) and, for the 32$\times$32 datasets, the same template with channel multipliers $(1,2,2,2)$, attention at 16$\times$16, and encoder multipliers $(1,2,2,2,2)$. We keep the official settings (batch size 128, Adam with learning rate $10^{-4}$, $T=1000$, $\epsilon$-prediction) and change only the semantic latent dimension (to $d_z$), the training length (to the number of training images used for InfoDiffusion), and the EMA decay of the autoencoder, from $0.9999$ to $0.9995$, because with $0.9999$ the averaged weights of our much shorter runs still contain a noticeable fraction of the random initialization. Unconditional samples use the official latent DDPM (the FFHQ-128 recipe, and the official CelebA-64 recipe for CelebA) and DDIM with $T=100$ for both the latent and the image. On FFHQ, the AE, VAE, $\beta$-VAE, and InfoVAE are re-trained with the released InfoDiffusion code (50 epochs, batch size 64) with no regularizer, a KL weight of $1$, a KL weight of $0.01$, and an MMD weight of $0.1$, respectively, and sampled with $\bz\sim\mathcal{N}(\mathbf{0},\mathbf{I})$; with a KL weight of $1$, the VAE collapses to nearly identical samples. The remaining entries of the AE, VAE, $\beta$-VAE, and InfoVAE, and the baselines in \Cref{tab:contrast1,tab:contrast2}, come from earlier runs with a separate training pipeline. For the re-trained baselines, $\pm$ is the standard deviation across the five folds (latent quality, TAD, and captured attributes), three evaluation subsamples (DCI), or three sets of $10{,}000$ samples (FID).

\paragraph{Evaluation} For latent quality, we fit a logistic regression classifier on the standardized encodings of training images with 5-fold cross-validation (for CelebA, the first 10{,}000 training images) and report accuracy (CIFAR10, FashionMNIST), the average AUROC over the 40 attributes (CelebA), or the average accuracy over age, gender, and eyeglasses (FFHQ). For FFHQ, we use the labels of the ffhq-features-dataset: age in integer years (treated as a class label), binary gender, and four eyeglasses categories; the 529 images without a detected face are excluded. We use the first 63{,}000 FFHQ images for training. TAD is computed on each of the five training folds of the first 10{,}000 CelebA images and averaged over folds. DCI is computed with gradient-boosted trees \citep{scikit-learn} fit on the encodings of 10{,}000 training images and evaluated on 5{,}000 test images. FID is computed with clean-FID \citep{parmar2022aliased} between 10{,}000 samples and the full training set. Unless stated otherwise, InfoDiffusion results are means $\pm$ one standard deviation over 3 training seeds, with each seed evaluated as above.

\begin{table}[ht]
    \caption{Dataset-specific settings for InfoDiffusion. The two dimensionalities of $\bz$ correspond to latent evaluation (`Eval.') and generation (`Gen.') models. Channels are the base widths of the decoder and the encoder. `Pool' is the operation applied to the final encoder feature map before the linear head; `Crop' indicates the random resized crops of the encoder input; `$t$' is the distribution of the training time step.}
    \label{tab:exp_details}
    \begin{small}
    \begin{sc}
    \begin{center}
    \setlength{\tabcolsep}{4.5pt}
    \begin{tabular}{lcccccccc}
    \toprule
        & Input size & \multicolumn{2}{c}{Dim. of $\bz$} & Channels & Pool & Crop & $t$ & Steps (epochs)\\
        & & Eval. & Gen. & dec.\,/\,enc. & & & & \\
        \midrule
         3DShapes & $3 \times 64 \times 64$ & 10 & -- & 40\,/\,48 & flatten & \ding{55} & uniform & 16{,}875 (5)\\
         FashionMNIST & $1 \times 32 \times 32$ & 32 & 256 & 40\,/\,48 & flatten & \ding{51} & logit-normal & 23{,}400 (50)\\
         CIFAR10 & $3 \times 32 \times 32$ & 32 & 256 & 80\,/\,96 & mean & \ding{51} & logit-normal & 19{,}500 (50)\\
         FFHQ & $3 \times 64 \times 64$ & 32 & 256 & 80\,/\,96 & flatten & \ding{51} & logit-normal & 24{,}600 (50)\\
         CelebA & $3 \times 64 \times 64$ & 32 & 256 & 80\,/\,96 & mean\,/\,flatten & \ding{51} & logit-normal & 63{,}600 (50)\\
         \bottomrule
    \end{tabular}
    \end{center}
    \end{sc}
    \end{small}
    \vspace{-0.2cm}
    \begin{center}\footnotesize On CelebA, the $d_z=32$ model uses mean pooling and the $d_z=256$ model uses flattening.\end{center}
\end{table}

\section{Additional Sampling Details}\label{app:sampling}
\subsection{Sampling from Prior}\label{app:sampling_prior}
To sample with $\bz\sim p(\bz)=\mathcal{N}(\mathbf{0},\mathbf{I})$, we draw $\bx_T\sim\mathcal{N}(\mathbf{0},\mathbf{I})$ and integrate the probability-flow ODE of the decoder from $t=1$ to $t=0$ with Heun's second-order method on a uniform grid of 50 steps (99 network evaluations, twice as many with guidance). Guidance on $\bz$ (\Cref{subsec:sampling}) requires a model trained with latent dropout; we use it only with the $d_z=256$ generative models.

\subsection{Sampling from Learned Prior}\label{app:sampling_learned}
To enable sampling from the learned prior, we train a latent diffusion model, analogous to the DiffAE approach \cite{preechakul2022diffusion}. We first train our InfoDiffusion model.
We then compute the latent representation $\bz$ for each image in a dataset using the trained $q_\phi(\bz\mid\bx)$ encoder.
Finally, a latent diffusion model is trained on these latent embeddings.
To generate using the learned latent, the decoder is conditioned on vectors $\bz$ sampled from the latent diffusion model.
Specifically, we standardize the encodings of the (un-augmented) training images per dimension and fit a velocity network $\bv_\psi(\bz_t, t)$ with the same flow-matching objective as the decoder (logit-normal $t$). The network is a residual MLP with width 1{,}024 and six blocks, each modulated by the time embedding through adaptive layer normalization (\Cref{fig:aux}, right). We train it for 30{,}000 steps with AdamW (learning rate $10^{-4}$, batch size 1{,}024) and keep an EMA of its weights. Latents are sampled with 100 Heun steps and de-standardized before decoding.

\section{Illustrations of Network Architecture}\label{app:arch}
Here we provide detailed illustrations of network architectures.
\Cref{fig:enc} shows the encoder, which is used for encoding original input images into low-dimensional latent embeddings.
\Cref{fig:res} shows the UNet decoder's architecture, which is used as the velocity prediction network in InfoDiffusion.
\Cref{fig:aux} shows the details of how we implement our Auxiliary Residual Block (left) and the MLP velocity network of the learned latent prior (right).

\begin{figure}[ht]
\centering
\begin{tikzpicture}[font=\footnotesize, >=Stealth,
  blk/.style={draw, rounded corners=2pt, minimum height=0.9cm, align=center, inner sep=3pt},
  res/.style={blk, fill=blue!8},
  op/.style={blk, fill=gray!12},
  io/.style={blk, fill=orange!15},
  lat/.style={blk, fill=green!15}]
  \node[io] (x) {$\bx_0$\\[-1pt]\scriptsize $H\times W$};
  \node[op, right=0.35cm of x] (crop) {random\\resized crop\\[-1pt]\scriptsize (train only)};
  \node[op, right=0.35cm of crop] (conv) {Conv\\$3\times3$};
  \node[res, right=0.35cm of conv] (l1) {$2\times$ Res\\[-1pt]\scriptsize $C$};
  \node[res, right=0.35cm of l1] (l2) {Down,\\$2\times$ Res\\[-1pt]\scriptsize $2C$};
  \node[res, right=0.35cm of l2] (l3) {Down,\\$2\times$ Res\\[-1pt]\scriptsize $2C$};
  \node[res, right=0.35cm of l3] (l4) {Down,\\$2\times$ Res\\[-1pt]\scriptsize $2C$};
  \node[res, right=0.35cm of l4] (mid) {Res+Attn,\\Res\\[-1pt]\scriptsize $\frac{H}{8}\times\frac{W}{8}$};
  \node[op, below=0.7cm of mid] (gn) {GN, SiLU};
  \node[op, left=0.35cm of gn] (pool) {mean pool\\\textit{or} flatten};
  \node[op, left=0.35cm of pool] (fc) {Linear};
  \node[lat, left=0.35cm of fc] (z) {$\bz$\\[-1pt]\scriptsize ($\bmu, \log\bsigma^2$ if $\zeta<1$;\\[-2pt]\scriptsize logits if discrete)};
  \foreach \a/\b in {x/crop,crop/conv,conv/l1,l1/l2,l2/l3,l3/l4,l4/mid,mid/gn,gn/pool,pool/fc,fc/z} \draw[->] (\a) -- (\b);
\end{tikzpicture}
\caption{The encoder of InfoDiffusion: the downsampling half of a U-Net (residual blocks without time conditioning; self-attention at $16\times16$ and $8\times8$) followed by a linear head on either the spatially averaged or the flattened final feature map. During training, the input is a random resized crop of $\bx_0$. For $\zeta<1$, the head outputs the mean and log-variance of a Gaussian $q_\phi(\bz\mid\bx_0)$; for discrete latents, its output is quantized with FSQ.}
\label{fig:enc}
\end{figure}

\begin{figure}[ht]
\centering
\begin{tikzpicture}[font=\footnotesize, >=Stealth,
  blk/.style={draw, rounded corners=2pt, minimum height=0.8cm, minimum width=1.55cm, align=center, inner sep=2pt},
  res/.style={blk, fill=blue!8},
  op/.style={blk, fill=gray!12},
  io/.style={blk, fill=orange!15, minimum width=1.0cm},
  emb/.style={blk, fill=green!15}]
  \node[io] (x) {$\bx_t$};
  \node[op, right=0.3cm of x] (conv) {Conv $3\times3$};
  \node[res, right=0.3cm of conv] (d1) {$2\times$ Res\\[-1pt]\scriptsize $k{=}1$};
  \node[res, right=0.3cm of d1] (d2) {Down, $2\times$ Res\\[-1pt]\scriptsize $k{=}2$};
  \node[res, right=0.3cm of d2] (d3) {Down, $2\times$ Res\\[-1pt]\scriptsize $k{=}2$};
  \node[res, right=0.3cm of d3] (d4) {Down, $2\times$ Res\\[-1pt]\scriptsize $k{=}2$};
  \node[res, below=1.1cm of d4] (m) {Res+Attn,\\Res};
  \node[res, left=0.3cm of m] (u4) {$3\times$ Res, Up\\[-1pt]\scriptsize $k{=}2$};
  \node[res, left=0.3cm of u4] (u3) {$3\times$ Res, Up\\[-1pt]\scriptsize $k{=}2$};
  \node[res, left=0.3cm of u3] (u2) {$3\times$ Res, Up\\[-1pt]\scriptsize $k{=}2$};
  \node[res, left=0.3cm of u2] (u1) {$3\times$ Res\\[-1pt]\scriptsize $k{=}1$};
  \node[op, left=0.3cm of u1] (out) {GN, SiLU,\\Conv $3\times3$};
  \node[io, left=0.3cm of out] (v) {$\bv_\theta$};
  \foreach \a/\b in {x/conv,conv/d1,d1/d2,d2/d3,d3/d4,d4/m,m/u4,u4/u3,u3/u2,u2/u1,u1/out,out/v} \draw[->] (\a) -- (\b);
  \foreach \a/\b in {d1/u1,d2/u2,d3/u3,d4/u4} \draw[->, densely dashed, gray] (\a) -- (\b);
  \node[emb, below=1.3cm of u1, minimum width=3.3cm] (te) {$t$: sinusoidal, MLP $\to \be_t$};
  \node[emb, right=0.5cm of te, minimum width=4.4cm] (ze) {$\bz$: MLP $\to \be_z$ \;\scriptsize(or null $\varnothing$)};
  \coordinate (busl) at ($(u1.south west)+(0,-0.45)$);
  \coordinate (busr) at ($(m.south east)+(0,-0.45)$);
  \draw[gray, densely dotted, thick] (busl) -- node[pos=0.97, below, font=\scriptsize, text=black]{to all residual blocks} (busr);
  \draw[->, gray] (te.north) -- (te.north |- busl);
  \draw[->, gray] (ze.north) -- (ze.north |- busl);
\end{tikzpicture}
\caption{The U-Net decoder of InfoDiffusion, conditioned on the time $t$ and the auxiliary variable $\bz$. The embeddings of $t$ and $\bz$ are passed to every residual block. With latent dropout, the embedding of $\bz$ is replaced by a learned null embedding $\varnothing$. $C$ is the base width and $k$ the channel multiplier of a level.}
\label{fig:res}
\end{figure}

\begin{figure}[ht]
\centering
\begin{tikzpicture}[font=\footnotesize, >=Stealth,
  blk/.style={draw, rounded corners=2pt, minimum height=0.55cm, minimum width=3.0cm, align=center, inner sep=2pt},
  op/.style={blk, fill=gray!12},
  mod/.style={blk, fill=green!15},
  io/.style={blk, fill=orange!15}]
  \node[io] (h) {$\bh$};
  \node[op, below=0.22cm of h] (a1) {GN, SiLU, Conv $3\times3$};
  \node[op, below=0.22cm of a1] (a2) {GroupNorm};
  \node[mod, below=0.22cm of a2] (a3) {$(1+\bs(\be_t))\,\cdot + \bb(\be_t)$};
  \node[mod, below=0.22cm of a3] (a4) {$(1+\bs(\be_z))\,\cdot + \bb(\be_z)$};
  \node[op, below=0.22cm of a4] (a5) {SiLU, Dropout};
  \node[op, below=0.22cm of a5] (a6) {Conv $3\times3$ \scriptsize(zero init)};
  \node[draw, circle, inner sep=1pt, below=0.22cm of a6] (plus) {$+$};
  \node[io, below=0.22cm of plus] (ho) {$\bh'$};
  \foreach \a/\b in {h/a1,a1/a2,a2/a3,a3/a4,a4/a5,a5/a6,a6/plus,plus/ho} \draw[->] (\a) -- (\b);
  \draw[->, gray] (h.east) -- ++(1.9,0) |- node[pos=0.25, right, align=left, font=\scriptsize]{skip\\(Conv $1\times1$\\if needed)} (plus.east);
  \node[left=0.25cm of a3, font=\scriptsize] {AGN with $t$};
  \node[left=0.25cm of a4, font=\scriptsize] {modulation by $\bz$};
  \begin{scope}[xshift=8.4cm]
  \node[io] (zt) {$\bz_t$ \scriptsize(standardized)};
  \node[op, below=0.22cm of zt] (b1) {Linear $\to 1024$};
  \node[op, below=0.45cm of b1] (b2) {LayerNorm};
  \node[mod, below=0.22cm of b2] (b3) {$(1+\bs(\be_t))\,\cdot + \bb(\be_t)$};
  \node[op, below=0.22cm of b3] (b4) {MLP ($4\times$ width, SiLU)};
  \node[mod, below=0.22cm of b4] (b5) {$\cdot\;\mathbf{g}(\be_t)$ \scriptsize(gate, zero init)};
  \node[draw, circle, inner sep=1pt, below=0.22cm of b5] (p2) {$+$};
  \node[op, below=0.4cm of p2] (b6) {LayerNorm, Linear};
  \node[io, below=0.22cm of b6] (vo) {$\bv_\psi(\bz_t, t)$};
  \foreach \a/\b in {zt/b1,b1/b2,b2/b3,b3/b4,b4/b5,b5/p2,p2/b6,b6/vo} \draw[->] (\a) -- (\b);
  \coordinate (r) at ($(b1.south)!0.5!(b2.north)$);
  \draw[->, gray] (r) -- ++(1.9,0) |- (p2.east);
  \node[draw, dashed, gray, rounded corners, fit=(b2)(b3)(b4)(b5)(p2), inner sep=4pt, label={[font=\scriptsize]left:$\times 6$}] {};
  \end{scope}
\end{tikzpicture}
\caption{\textit{Left}: the auxiliary residual block. The normalized features are modulated first by the time embedding (AGN) and then by the embedding of $\bz$; the second convolution is zero-initialized. \textit{Right}: the MLP velocity network of the learned latent prior, with adaptive layer normalization (scale, shift, and gate predicted from the time embedding) in each of its six blocks.}
\label{fig:aux}
\end{figure}

\section{Ablation: Different Approach for Conditioning on $\bz$}\label{app:ablation}
We perform an ablation in which we condition on $\bz$ only in the bottleneck layer of the UNet, denoted as `Ours w/ bott. only' in \Cref{tab:arch1} and \Cref{tab:arch2}, as opposed to conditioning at all layers.
Conditioning on $\bz$ at all layers (`Ours') gives more informative latents (measured by the average accuracy/AUROC in predicting attributes from latents) on all four datasets and lower FID on FashionMNIST and CIFAR10. Bottleneck-only conditioning gives slightly lower FID on FFHQ and CelebA ($12.3$ vs.\ $12.5$ and $7.6$ vs.\ $7.7$) and a higher TAD score on CelebA ($0.654$ vs.\ $0.555$), with the same number of captured attributes.

\begin{table*}[ht!]
\setlength{\tabcolsep}{2pt}
    \caption{Comparison between two modeling choices: conditioning on $\bz$ at just the bottleneck layer of the UNet (denoted as `Ours w/ bott. only') versus at all layers.
    `Latent quality' is measured as classification accuracy/AUROC for logistic regression classifiers trained on the auxiliary latent vector $\bz.$
    Both variants use $\lambda=0.1$, $\zeta=0.9$ and a single training seed; latent quality is measured with $d_z=32$ models and FID with $d_z=256$ models, using the learned latent prior without guidance.
    Darkly shaded cells indicate best results.
    }
    \label{tab:arch1}
    \begin{center}
    \begin{small}
    \begin{sc}
    \begin{tabular}{p{1.8cm}l cc cc  cc  cc}
    \toprule
    &\multicolumn{2}{c}{FashionMNIST}&\multicolumn{2}{c}{CIFAR10}&\multicolumn{2}{c}{FFHQ}&\multicolumn{2}{c}{CelebA}\\
    \midrule
    & \makecell{Latent \\Qual. $\uparrow$} & FID $\downarrow$ & \makecell{Latent \\Qual. $\uparrow$} & FID $\downarrow$ &  \makecell{Latent \\Qual. $\uparrow$} & FID $\downarrow$ & \makecell{Latent \\Qual. $\uparrow$} & FID $\downarrow$\\
    \midrule
    \makecell{Ours (with \\bott. only)} & $0.816$ & $8.0$ & $0.355$ & $26.6$ & $0.625$ & \firstplace$12.3$ & $0.817$ & \firstplace$7.6$\\
    Ours & \firstplace$0.852$ & \firstplace$7.7$ & \firstplace$0.457$ & \firstplace$25.7$ & \firstplace$0.629$ & $12.5$ & \firstplace$0.844$ & $7.7$\\
    \bottomrule
    \end{tabular}
    \end{sc}
    \end{small}
    \end{center}
\end{table*}

\begin{table}[ht!]
\setlength{\tabcolsep}{2pt}
    \caption{Comparison between two modeling choices: conditioning on $\bz$ at just the bottleneck layer of the UNet (denoted as `Ours w/ bott. only') versus at all layers.
    `Attrs.' counts the number of ``captured'' attributes when calculating the TAD score.
    Single training seed ($\lambda=0.1$, $\zeta=0.9$, $d_z=32$).
    Darkly shaded cells indicate best result.}
    \label{tab:arch2}
    \begin{center}
    \begin{small}
    \begin{sc}
    \begin{tabular}{p{4cm}l c@{\hskip 1em}c }
    \toprule
    CelebA & TAD $\uparrow$ & ATTRS $\uparrow$ \\
    \midrule
    Ours (with bott. only) & \firstplace$0.654$ & \firstplace$6.0$ \\
    Ours & $0.555$ & \firstplace$6.0$ \\
    \bottomrule
    \end{tabular}
    \end{sc}
    \end{small}
    \end{center}
\end{table}

\section{Discrete Latents}\label{app:discrete}
Often factors of variation can be described by categorical or binary variables.
For example, the CelebA dataset contains binary annotations for each image indicating the presence or absence of certain attributes, e.g., facial hair.
For this and similar datasets, it might be more appropriate to model the auxiliary latent variables as categorical, e.g., a vector of Bernoulli variables, rather than the typical continuous Gaussian distribution.

\paragraph{Finite scalar quantization} Our discrete latents use FSQ \citep{mentzer2024finite} (\Cref{subsec:discrete}). The encoder is that of the CelebA latent model in \Cref{tab:exp_details} (mean pooling, $d_z=32$); its output $\bh\in\mathbb{R}^{32}$ is mapped to $z_i = \mathrm{round}(2\tanh(h_i))/2 \in \{-1, -\tfrac{1}{2}, 0, \tfrac{1}{2}, 1\}$ ($L=5$), with straight-through gradients. The prior is uniform over the $5^{32}$ codes, there is no auxiliary loss, and all other settings, including the absence of latent dropout, are those of the CelebA latent model. Latent quality and TAD are computed on the quantized codes with the protocol of \Cref{sec:experiments}. Over three seeds, FSQ latents reach an average AUROC of $0.849 \pm 0.003$, a TAD score of $0.370 \pm 0.240$, and $5.6 \pm 0.9$ captured attributes, on par with the continuous latents of the main configuration (\Cref{tab:discrete}). Almost every training image receives its own code ($9{,}963$ distinct codes among the first $10{,}000$ images for the first seed).

\paragraph{Relaxed Bernoulli latents} A natural alternative for binary attributes is a Relaxed-Bernoulli distribution, which is derived from the Gumbel-Softmax trick \cite{jang2016categorical} for categorical variables, an extension of the reparameterization trick \cite{kingma2013auto} to categorical distributions.
This defines a ``soft" or ``smooth" version of the Bernoulli distribution, which enables gradient based optimization.
The Gumbel-Softmax distribution, also known as the Concrete distribution \citep{maddison2017concrete}, is a way of drawing samples $z$ from a categorical distribution with $k$ classes, but with a differentiable function. If $\pi_i$ represents the probability of class $i$, the Gumbel-Softmax distribution is defined as:
\begin{equation}\label{eq:gumbel}
z_i = \frac{\exp((\log(\pi_i) + g_i)/\tau)}{\sum_{j=1}^{k}\exp((\log(\pi_j) + g_j)/\tau)}
\end{equation}
where $g_i$ are i.i.d. samples drawn from the Gumbel(0, 1) distribution and $\tau$ is the temperature parameter. The temperature parameter controls the randomness of samples. As $\tau \rightarrow 0$, samples become one-hot encoded (more deterministic), and as $\tau \rightarrow \infty$, samples approach a uniform distribution.
For the Relaxed-Bernoulli, we have $k = 2,$ in \Cref{eq:gumbel}.
We tested this relaxation with an encoder that outputs one logit $\ell_i$ per latent dimension ($d_z=32$), training samples $z_i = \sigma((\ell_i + g_i)/\tau)$ with logistic noise $g_i$, a temperature that decreases linearly from $1$ to $0.5$, a Bernoulli$(1/2)$ prior, and binary codes $z_i = \mathbb{1}[\ell_i>0]$ at test time. We matched the aggregate posterior to the prior with MMD (weight $0.01$), used a per-sample KL term to the prior (weight $0.1$), or used no regularizer, and we also tried a straight-through version with binary codes in the forward pass. All of these variants learn codes that carry little attribute information (AUROC $0.57$--$0.61$, no captured attributes; \Cref{tab:discrete}).

\paragraph{Binary codes without relaxation} Removing the relaxation noise helps but does not close the gap: deterministic binary codes $z_i=\mathrm{sign}(h_i)$ with straight-through gradients and MMD to the Bernoulli$(1/2)$ prior (weight $0.01$) reach an AUROC of $0.712 \pm 0.014$, within the range of the continuous-latent baselines, but capture almost no attributes. Lookup-free quantization \citep{yu2024language}, which adds an entropy term that encourages diverse codes, collapsed to a single code with its commitment loss (weight $0.25$) and reached an AUROC of $0.52$ without it.

\paragraph{Diagnosis} In the variants that fail, the decoder largely ignores $\bz$: decoding with $\bz$ and a random $\bx_T$ reconstructs test images much worse than with FSQ or continuous latents (\Cref{tab:discrete}). For the relaxed variants, linear probes on the encoder logits are only slightly better than on the binary codes (AUROC $0.59$--$0.66$), so the information is not learned rather than lost by thresholding. Early in training, the relaxed samples are dominated by noise, and with a decoder that predicts well from $\bx_t$ alone, the decoder learns to ignore $\bz$; once the logits saturate, the encoder receives little gradient and its codes stay uninformative (only $618$--$2{,}870$ distinct codes among $10{,}000$ images). FSQ codes are deterministic, like the binary sign codes, but have five levels per dimension; in our experiments, only this finer quantization matched the continuous latents.

\begin{table}[ht]
    \caption{Discrete latents on CelebA ($d_z=32$). Latent quality (average AUROC), TAD, and captured attributes follow the protocol of \Cref{tab:disentangle}; `$\bz$-only MSE' is the mean squared error (images in $[-1,1]$) of decoding $1{,}000$ test images from their latents and a random $\bx_T$. Means $\pm$ standard deviations are over three seeds; other rows use one seed.}
    \label{tab:discrete}
    \begin{center}
    \begin{small}
    \begin{tabular}{lccccc}
    \toprule
    Latent & Seeds & Latent Qual.\,$\uparrow$ & TAD\,$\uparrow$ & Attrs\,$\uparrow$ & $\bz$-only MSE\,$\downarrow$ \\
    \midrule
    Continuous, $(\lambda,\zeta)=(0.1,0.9)$ & 3 & $0.844 \pm 0.005$ & $0.439 \pm 0.134$ & $5.7 \pm 0.6$ & $0.221$ \\
    \midrule
    FSQ ($L=5$) & 3 & $0.849 \pm 0.003$ & $0.370 \pm 0.240$ & $5.6 \pm 0.9$ & $0.231$ \\
    Binary $\mathrm{sign}(h)$, MMD $0.01$ & 3 & $0.712 \pm 0.014$ & $0.030 \pm 0.026$ & $0.5 \pm 0.4$ & $0.337$ \\
    Binary $\mathrm{sign}(h)$, no regularizer & 1 & $0.687$ & $0.000$ & $0.0$ & $0.365$ \\
    Relaxed Bernoulli, MMD $0.01$ & 3 & $0.574 \pm 0.011$ & $0.000$ & $0.0$ & $0.588$ \\
    Relaxed Bernoulli, KL $0.1$ & 3 & $0.613 \pm 0.008$ & $0.000$ & $0.0$ & $0.437$ \\
    Relaxed Bernoulli, no regularizer & 1 & $0.596$ & $0.026$ & $1.0$ & $0.553$ \\
    Relaxed Bernoulli, straight-through, KL $0.1$ & 1 & $0.603$ & $0.007$ & $0.8$ & $0.465$ \\
    Relaxed Bernoulli, straight-through & 1 & $0.579$ & $0.000$ & $0.0$ & $0.582$ \\
    LFQ, with commitment & 1 & $0.500$ & $0.000$ & $0.0$ & $0.603$ \\
    LFQ, without commitment & 1 & $0.520$ & $0.000$ & $0.0$ & $0.492$ \\
    \bottomrule
    \end{tabular}
    \end{small}
    \end{center}
\end{table}

\section{Regularization Coefficients}\label{app:coeffs}
In \Cref{app:tab:class} and \Cref{app:tab:disentangle}, we copy results from \Cref{tab:class} and \Cref{tab:disentangle}, respectively, and add results for other choices of the $\lambda$ and $\zeta$ regularization coefficients.
In our implementation, the coefficients enter the loss as $(1-\zeta)$ for the per-sample KL term and $(\lambda+\zeta-1)$ for the MMD term; e.g., $(\lambda,\zeta)=(0.1,0.9)$ uses only the KL term with weight $0.1$, and $(0.1,0.95)$ uses weights $0.05$ for both terms. As described in \Cref{sec:5}, the per-sample KL term is divided by the data dimension, so that $(1-\zeta)$ is its weight relative to the per-pixel denoising loss, as in a per-dimension ELBO.
All settings use a single training seed, except $(\lambda,\zeta)=(0.1,0.9)$ and $(0.01,1)$, and the latent models with $(0,1)$, which use three seeds; for these, $\pm$ is the standard deviation over seeds. For single-seed settings, $\pm$ is the standard deviation across the five folds (latent quality, TAD, and captured attributes), three evaluation subsamples (DCI), or three sets of $10{,}000$ samples from the same model (FID); these error bars do not include the variation across training runs. FID is computed without guidance. $(\lambda,\zeta)=(0.1,0.9)$ gives the best or second-best FID with both samplers on all four natural image datasets and the most disentangled CelebA latents. With $\zeta=1$, the prior regularizer trades latent information for prior matching: increasing $\lambda$ from $0$ to $0.1$ lowers latent quality on CIFAR10, FFHQ, and CelebA, and $\lambda=0.1$ removes the attribute information from the CelebA latents. $\lambda=0$, which corresponds to DiffAE, gives the most informative latents on CIFAR10, FashionMNIST, and CelebA, but poor samples from $p(\bz)$ on CIFAR10, FFHQ, and CelebA. $(0.1,0.95)$ collapses the scale of the latents on CIFAR10, FFHQ, and CelebA (average standard deviation below $0.2$) and loses much of their information, while giving the highest DCI on 3DShapes ($0.429$).
\begin{table*}[ht!]
\setlength{\tabcolsep}{2pt}
    \caption{Latent quality, as measured by classification accuracies for logistic regression classifiers trained on the auxiliary latent vector $\bz,$ and FID.
    We report mean $\pm$ one standard deviation (\Cref{app:coeffs}).
    Darkly shaded cells indicate the best while lightly shaded cells indicate the second best. InfoDiffusion FIDs use $w=1$.
    }
    \label{app:tab:class}
    \begin{center}
    \begin{small}
    \begin{sc}
    \begin{tabular}{l @{\hskip 1.5em} cc @{\hskip 1.5em} cc @{\hskip 1.5em} cc}
    \toprule
    &\multicolumn{2}{c}{FashionMNIST}&\multicolumn{2}{c}{CIFAR10}&\multicolumn{2}{c}{FFHQ}\\
    \midrule
    & \makecell{Latent \\Qual. $\uparrow$} & FID $\downarrow$ & \makecell{Latent \\Qual. $\uparrow$} & FID $\downarrow$ &  \makecell{Latent \\Qual. $\uparrow$} & FID $\downarrow$\\
    \midrule
    AE & $0.819 \pm 0.003$ & $62.9 \pm 2.1$ & $0.336 \pm 0.005$ & $169.4 \pm 2.4$ & $0.532 \pm 0.002$ & $316.7 \pm 0.4$ \\
    VAE & $0.796 \pm 0.002$ & $63.4 \pm 1.6$ & $0.342 \pm 0.004$ & $177.2 \pm 3.2$ & $0.488 \pm 0.002$ & $307.7 \pm 0.1$ \\
    beta-VAE & $0.779 \pm 0.004$ & $66.9 \pm 1.8$ & $0.253 \pm 0.003$ & $183.3 \pm 3.1$ & $0.510 \pm 0.001$ & $222.6 \pm 0.2$ \\
    InfoVAE & $0.807 \pm 0.003$ & $55.0 \pm 1.7$ & $0.357 \pm 0.005$ & $160.7 \pm 2.5$ & $0.545 \pm 0.001$ & $195.2 \pm 0.5$ \\
    DiffAE & $0.822 \pm 0.006$ & \firstplace$5.6 \pm 0.1$ & $0.442 \pm 0.004$ & $34.1 \pm 0.1$ & $0.565 \pm 0.001$ & \firstplace$12.1 \pm 0.1$ \\
    \midrule
    InfoDiffusion $(\lambda,\zeta)=(0.1, 0.9)$ & \secondplace$0.856 \pm 0.004$ & $22.5 \pm 0.6$ & $0.461 \pm 0.003$ & $36.2 \pm 1.1$ & \firstplace$0.628 \pm 0.002$ & $19.8 \pm 0.4$ \\
    ~~ w/Learned Latent &  & \secondplace$8.1 \pm 0.4$ &  & \firstplace$25.7 \pm 0.4$ &  & \secondplace$13.1 \pm 0.5$ \\
    InfoDiffusion $(\lambda,\zeta)=(0.1, 0.95)$ & $0.848 \pm 0.002$ & $27.2 \pm 0.3$ & $0.359 \pm 0.003$ & $37.4 \pm 0.2$ & $0.497 \pm 0.001$ & $19.4 \pm 0.2$ \\
    ~~ w/Learned Latent &  & \secondplace$8.1 \pm 0.1$ &  & $26.7 \pm 0.3$ &  & $13.2 \pm 0.2$ \\
    InfoDiffusion $(\lambda,\zeta)=(0.1, 1)$ & $0.852 \pm 0.004$ & $41.7 \pm 0.5$ & $0.451 \pm 0.002$ & $41.2 \pm 0.2$ & $0.545 \pm 0.001$ & $24.0 \pm 0.2$ \\
    ~~ w/Learned Latent &  & $8.5 \pm 0.0$ &  & $27.0 \pm 0.3$ &  & $15.6 \pm 0.1$ \\
    InfoDiffusion $(\lambda,\zeta)=(0.01, 1)$ & \firstplace$0.862 \pm 0.002$ & $38.1 \pm 2.6$ & \secondplace$0.512 \pm 0.004$ & $64.0 \pm 0.6$ & $0.606 \pm 0.005$ & $41.5 \pm 1.6$ \\
    ~~ w/Learned Latent &  & $8.8 \pm 0.4$ &  & \secondplace$26.1 \pm 0.3$ &  & $15.1 \pm 0.2$ \\
    InfoDiffusion $(\lambda,\zeta)=(0, 1)$ & \firstplace$0.862 \pm 0.003$ & $14.9 \pm 0.1$ & \firstplace$0.516 \pm 0.006$ & $60.8 \pm 0.1$ & \secondplace$0.624 \pm 0.004$ & $51.0 \pm 0.6$ \\
    ~~ w/Learned Latent &  & $9.1 \pm 0.2$ &  & $27.6 \pm 0.2$ &  & $15.3 \pm 0.2$ \\
    \bottomrule
    \end{tabular}
    \end{sc}
    \end{small}
    \end{center}
    \label{tab:app_class}
\end{table*}

\begin{table*}[ht!]
\setlength{\tabcolsep}{2pt} %
    \caption{Disentanglement and latent quality metrics and FID.
    For 3DShapes, we check the image quality manually and label the models which generate high-quality images with check marks (`Image Qual.').
    The visualization of the samples is shown in \Cref{fig:3dshapes_visual} in the \Cref{app:3dshapes}; for InfoDiffusion, image quality is judged from 16 samples drawn with $\bz\sim p(\bz)$ (first row of each setting) or from the learned latent prior (second row).
    For CelebA, `Attrs.' counts the number of ``captured'' attributes when calculating the TAD score.
    `Latent Quality' is measured as AUROC scores averaged across attributes for logistic regression classifiers trained on the auxiliary latent vector $\bz$.
    We report means $\pm$ one standard deviation for quantitative metrics (\Cref{app:coeffs}).
    Darkly shaded cells indicate the best while lightly shaded cells indicate the second best. InfoDiffusion FIDs use $w=1$.}
    \newcommand{\greencheck}{{\bf \color{OliveGreen}\checkmark}}

    \newcommand*\colourcheck[1]{%
      \expandafter\newcommand\csname #1check\endcsname{\textcolor{#1}{\ding{52}}}%
    }
    \definecolor{bloodred}{HTML}{B00000}
    \definecolor{cautionyellow}{HTML}{EED202}
    \newcommand*\colourxmark[1]{%
      \expandafter\newcommand\csname #1xmark\endcsname{\textcolor{#1}{\ding{54}}}%
    }
    \newcommand*\colourcheckodd[1]{%
      \expandafter\newcommand\csname #1checkodd\endcsname{\textcolor{#1}{\ding{51}}}%
    }
    \colourcheckodd{cautionyellow}
    \colourcheck{cautionyellow}
    \colourcheck{OliveGreen}
    \colourxmark{bloodred}

    \newcommand{\ourxmark}{\bloodredxmark}%
    \newcommand{\ourcheckmark}{\OliveGreencheck}
    \label{app:tab:disentangle}
    \begin{center}
    \begin{small}
    \begin{sc}
    \begin{tabular}{l @{\hskip 1.2em} cc @{\hskip 1.2em} cccc}
    \toprule
    & \multicolumn{2}{c}{3DShapes}
    & \multicolumn{4}{c}{CelebA}\\
    \midrule
    & DCI $\uparrow$ & Image Qual. & TAD$\uparrow$ & Attrs$\uparrow$ & Latent Qual. $\uparrow$ & FID $\downarrow$ \\
    \midrule
    AE & $0.219 \pm 0.001$ & \ourxmark & $0.042 \pm 0.004$ & $1.0 \pm 0.0$ & $0.759 \pm 0.003$ & $90.4 \pm 1.8$ \\
    VAE & $0.276 \pm 0.001$ & \ourxmark & $0.000 \pm 0.000$ & $0.0 \pm 0.0$ & $0.770 \pm 0.002$ & $94.3 \pm 2.8$ \\
    beta-VAE & $0.281 \pm 0.001$ & \ourxmark & $0.088 \pm 0.051$ & $1.6 \pm 0.8$ & $0.699 \pm 0.001$ & $99.8 \pm 2.4$ \\
    InfoVAE & $0.134 \pm 0.001$ & \ourxmark & $0.000 \pm 0.000$ & $0.0 \pm 0.0$ & $0.757 \pm 0.003$ & $77.8 \pm 1.6$ \\
    DiffAE & $0.177 \pm 0.001$ & \ourcheckmark & $0.114 \pm 0.033$ & $2.8 \pm 0.4$ & $0.817 \pm 0.001$ & $8.4 \pm 0.1$ \\
    \midrule
    InfoDiffusion $(\lambda,\zeta)=(0.1, 0.9)$ & \secondplace$0.296 \pm 0.057$ & \ourxmark & \firstplace$0.439 \pm 0.134$ & \firstplace$5.7 \pm 0.6$ & $0.844 \pm 0.005$ & $12.1 \pm 0.2$ \\
    ~~ w/Learned Latent &  & \ourcheckmark &  &  &  & \firstplace$7.6 \pm 0.2$ \\
    InfoDiffusion $(\lambda,\zeta)=(0.1, 0.95)$ & \firstplace$0.429 \pm 0.001$ & \ourcheckmark & $0.000 \pm 0.000$ & $0.0 \pm 0.0$ & $0.632 \pm 0.004$ & $16.0 \pm 0.1$ \\
    ~~ w/Learned Latent &  & \ourcheckmark &  &  &  & \secondplace$8.1 \pm 0.2$ \\
    InfoDiffusion $(\lambda,\zeta)=(0.1, 1)$ & $0.166 \pm 0.000$ & \ourcheckmark & $0.000 \pm 0.000$ & $0.0 \pm 0.0$ & $0.606 \pm 0.001$ & $27.7 \pm 0.1$ \\
    ~~ w/Learned Latent &  & \ourcheckmark &  &  &  & $9.1 \pm 0.1$ \\
    InfoDiffusion $(\lambda,\zeta)=(0.01, 1)$ & $0.269 \pm 0.041$ & \ourcheckmark & $0.140 \pm 0.076$ & $3.5 \pm 0.5$ & \secondplace$0.867 \pm 0.002$ & $44.7 \pm 2.9$ \\
    ~~ w/Learned Latent &  & \ourcheckmark &  &  &  & $8.9 \pm 0.2$ \\
    InfoDiffusion $(\lambda,\zeta)=(0, 1)$ & $0.251 \pm 0.044$ & \ourxmark & \secondplace$0.299 \pm 0.105$ & \secondplace$5.5 \pm 0.6$ & \firstplace$0.875 \pm 0.000$ & $108.1 \pm 0.6$ \\
    ~~ w/Learned Latent &  & \ourcheckmark &  &  &  & $8.7 \pm 0.1$ \\
    \bottomrule
    \end{tabular}
    \end{sc}
    \end{small}
    \end{center}
\end{table*}

\section{Qualitative Figures on 3DShapes}\label{app:3dshapes}
In \Cref{fig:3dshapes_visual}, we show samples from unconditional generation of the images by different models.
The images generated by diffusion models (DiffAE and InfoDiffusion) are of high quality, with clear shapes and boundaries of the objects and backgrounds.
The images generated by VAE-based models suffer from distorted shapes and blended objects.
The results show that images generated by diffusion models are of higher quality than those generated by the VAE-based methods.
Of note, with latents drawn from the learned latent prior (\Cref{tab:disentangle}), our model is able to maintain high-quality image generation while attaining the best disentanglement metrics compared to the baseline models. Uncurated samples of our main configuration on the other datasets are shown in \Cref{app:samples}.

\begin{figure*}[ht]
    \centering
\setlength{\itemwidth}{0.06\linewidth}
\setlength{\tabcolsep}{2pt} %
\begin{tabular}{cccccccc}
\includegraphics[width=\itemwidth]{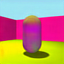} & \includegraphics[width=\itemwidth]{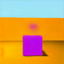} & \includegraphics[width=\itemwidth]{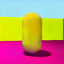} & \includegraphics[width=\itemwidth]{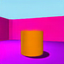} & \includegraphics[width=\itemwidth]{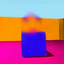} & \includegraphics[width=\itemwidth]{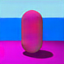} & \includegraphics[width=\itemwidth]{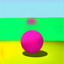} & \includegraphics[width=\itemwidth]{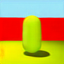} \\
\multicolumn{8}{c}{\small{(a) VAE}} \\
\includegraphics[width=\itemwidth]{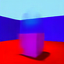} & \includegraphics[width=\itemwidth]{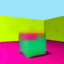} & \includegraphics[width=\itemwidth]{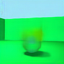} & \includegraphics[width=\itemwidth]{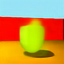} & \includegraphics[width=\itemwidth]{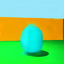} & \includegraphics[width=\itemwidth]{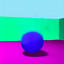} & \includegraphics[width=\itemwidth]{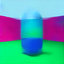} & \includegraphics[width=\itemwidth]{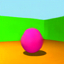} \\
\multicolumn{8}{c}{\small{(b) InfoVAE}} \\
\includegraphics[width=\itemwidth]{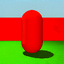} & \includegraphics[width=\itemwidth]{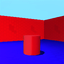} & \includegraphics[width=\itemwidth]{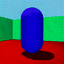} & \includegraphics[width=\itemwidth]{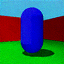} & \includegraphics[width=\itemwidth]{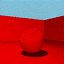} & \includegraphics[width=\itemwidth]{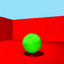} & \includegraphics[width=\itemwidth]{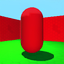} & \includegraphics[width=\itemwidth]{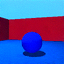} \\
\multicolumn{8}{c}{\small{(c) DiffAE}} \\
\includegraphics[width=\itemwidth]{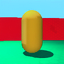} & \includegraphics[width=\itemwidth]{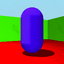} & \includegraphics[width=\itemwidth]{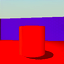} & \includegraphics[width=\itemwidth]{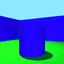} & \includegraphics[width=\itemwidth]{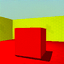} & \includegraphics[width=\itemwidth]{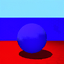} & \includegraphics[width=\itemwidth]{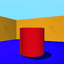} & \includegraphics[width=\itemwidth]{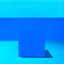} \\
\multicolumn{8}{c}{\small{(d) InfoDiff}} \\
\end{tabular}
\caption{Visualization of image samples from unconditional generation by VAE-based models (a-b) and diffusion models (c-d).}
\label{fig:3dshapes_visual}
\end{figure*}

\section{Uncurated Samples}\label{app:samples}
\Cref{fig:samples_learned,fig:samples_prior} show uncurated samples from the main configuration, $(\lambda,\zeta)=(0.1,0.9)$, trained with the first seed: each panel contains the first 32 (\Cref{fig:samples_learned}) or 64 (\Cref{fig:samples_prior}) samples for a fixed random seed, without any selection. For FashionMNIST, CIFAR10, CelebA, and FFHQ, we use the $d_z=256$ generative models with guidance on $\bz$ ($w=2$) and draw $\bz$ either from the learned latent prior (\Cref{fig:samples_learned}) or from $p(\bz)=\mathcal{N}(\mathbf{0},\mathbf{I})$ (\Cref{fig:samples_prior}); the corresponding FIDs are reported in \Cref{tab:class,tab:disentangle}. Samples with $\bz\sim p(\bz)$ contain more distorted faces and less recognizable objects, in line with their higher FIDs. All samples use 50 Heun steps (\Cref{app:sampling}); samples on 3DShapes are shown in \Cref{fig:3dshapes_visual}.

\begin{figure}[ht]
    \centering
    \subfigure[FashionMNIST]{\includegraphics[width=0.48\textwidth]{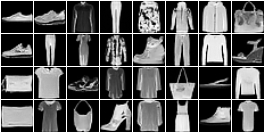}}\hfill
    \subfigure[CIFAR10]{\includegraphics[width=0.48\textwidth]{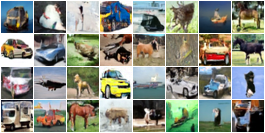}}\\
    \subfigure[CelebA]{\includegraphics[width=0.48\textwidth]{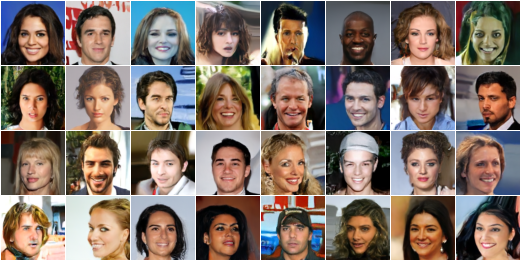}}\hfill
    \subfigure[FFHQ]{\includegraphics[width=0.48\textwidth]{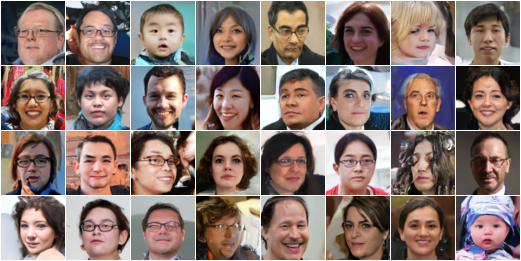}}
    \caption{Uncurated samples of InfoDiffusion ($\lambda=0.1$, $\zeta=0.9$, $d_z=256$) with $\bz$ drawn from the learned latent prior and guidance $w=2$.}
    \label{fig:samples_learned}
\end{figure}

\begin{figure}[ht]
    \centering
    \subfigure[FashionMNIST]{\includegraphics[width=0.48\textwidth]{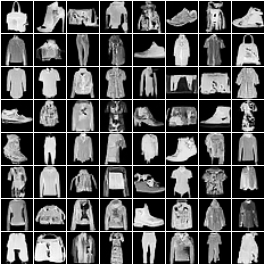}}\hfill
    \subfigure[CIFAR10]{\includegraphics[width=0.48\textwidth]{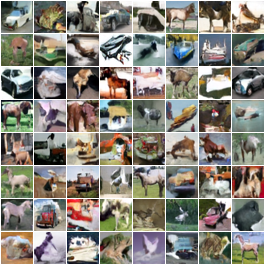}}\\
    \subfigure[CelebA]{\includegraphics[width=0.48\textwidth]{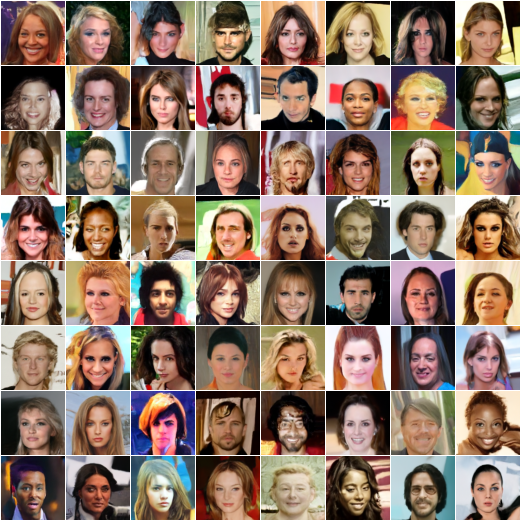}}\hfill
    \subfigure[FFHQ]{\includegraphics[width=0.48\textwidth]{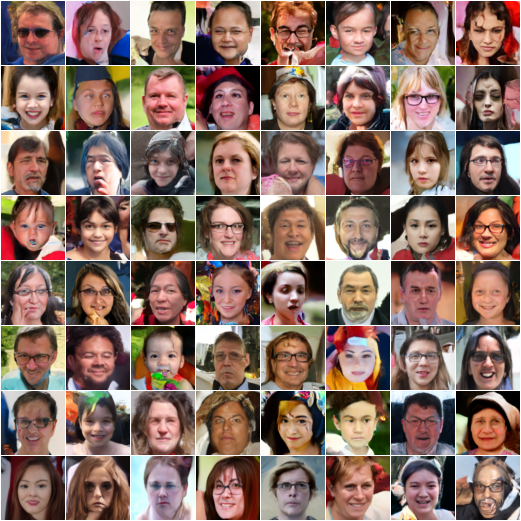}}
    \caption{Uncurated samples of InfoDiffusion ($\lambda=0.1$, $\zeta=0.9$, $d_z=256$) with $\bz\sim p(\bz)$ and guidance $w=2$.}
    \label{fig:samples_prior}
\end{figure}

\newpage
\section{Assets}\label{app:assets}
Below we list the libraries and datasets that we use in our experiments with their corresponding citations and licenses (in parentheses).
\paragraph{Libraries} We use the following open-source libraries: pytorch \cite{paszke2019pytorch} (license: BSD),
and scikit-learn \cite{scikit-learn} (BSD 3-Clause).

\paragraph{Datasets}
Our experimental section uses the following datasets: FashionMNIST~\cite{xiao2017fashion} (MIT), CIFAR10~\cite{krizhevsky2009learning} (MIT), FFHQ~\cite{karras2019style} (Creative Commons BY-NC-SA 4.0), CelebA \cite{liu2015faceattributes}, and 3DShapes \cite{3dshapes18} (Apache 2.0).

\end{document}